\documentclass[letterpaper]{article} % DO NOT CHANGE THIS
\usepackage{aaai2026}  % DO NOT CHANGE THIS
\usepackage{times}  % DO NOT CHANGE THIS
\usepackage{helvet}  % DO NOT CHANGE THIS
\usepackage{courier}  % DO NOT CHANGE THIS
\usepackage[hyphens]{url}  % DO NOT CHANGE THIS
\usepackage{graphicx} % DO NOT CHANGE THIS
\usepackage{natbib}  % DO NOT CHANGE THIS AND DO NOT ADD ANY OPTIONS TO IT
\usepackage{caption} % DO NOT CHANGE THIS AND DO NOT ADD ANY OPTIONS TO IT
\usepackage{algorithm}
\usepackage{algorithmic}

\usepackage{newfloat}
\usepackage{listings}
\DeclareCaptionStyle{ruled}{labelfont=normalfont,labelsep=colon,strut=off} % DO NOT CHANGE THIS
\floatstyle{ruled}
\newfloat{listing}{tb}{lst}{}
\floatname{listing}{Listing}
\usepackage[table,dvipsnames]{xcolor}         % colors
\definecolor{linkColor}{rgb}{0.18,0.39,0.62} % Dark Skyblue.
\definecolor{HeaderBlue}{RGB}{234,242,255}
\definecolor{red}{RGB}{220, 0, 0} 
\definecolor{blue}{RGB}{0, 60, 200}

\usepackage{amsfonts}
\usepackage{amsmath}
\usepackage{microtype}
\usepackage{booktabs} % for professional tables
\usepackage{multirow}
\usepackage{pifont}
\usepackage{stmaryrd}

\usepackage{hhline}
\usepackage{arydshln}

\newcommand{\eg}{{\it e.g., }}
\newcommand{\ie}{{\it i.e., }}

\definecolor{brickred}{rgb}{0.8, 0.25, 0.33}
\definecolor{brickgreen}{rgb}{0.25, 0.8, 0.33}
\newcommand{\cm}{\textcolor{brickgreen}{\ding{51}}}%
\newcommand{\xm}{\textcolor{brickred}{\ding{55}}}%
\usepackage{color}
\definecolor{brickred}{rgb}{0.8, 0.25, 0.33}
\definecolor{brickred2}{rgb}{0.25, 0.8, 0.33}
\usepackage{graphicx}

\newcommand{\ttabref}[1]{Tab.~\ref{#1}}
\newcommand{\ffigref}[1]{Fig.~\ref{#1}}

\newcommand{\eeqref}[1]{Eq.~(\ref{#1})}

\newcommand{\ssim}{\raisebox{0.5ex}{\texttildelow}}

\newcommand{\framework}{DegFlow}

\title{Continuous Degradation Modeling via Latent Flow Matching \\ for Real-World Super-Resolution}
\author{
    Hyeonjae Kim\textsuperscript{\rm 1}\equalcontrib, 
    Dongjin Kim\textsuperscript{\rm 1}\equalcontrib,
    Eugene Jin\textsuperscript{\rm 2}, 
    Tae Hyun Kim\textsuperscript{\rm 1}\thanks{Corresponding Author.}
}
\affiliations{
    \textsuperscript{\rm 1}Dept. of Computer Science, Hanyang University, Seoul, South Korea
    \\ \textsuperscript{\rm 2}Dept. of Artificial Intelligence Application, Hanyang University, Seoul, South Korea
    \{khj1112, dongjinkim, eugenebori, taehyunkim\}@hanyang.ac.kr
}

\begin{document}

\maketitle

\begin{abstract}
While deep learning-based super-resolution (SR) methods have shown impressive outcomes with synthetic degradation scenarios such as bicubic downsampling, they frequently struggle to perform well on real-world images that feature complex, nonlinear degradations like noise, blur, and compression artifacts. Recent efforts to address this issue have involved the painstaking compilation of real low-resolution (LR) and high-resolution (HR) image pairs, usually limited to several specific downscaling factors. To address these challenges, our work introduces a novel framework capable of synthesizing authentic LR images from a single HR image by leveraging the latent degradation space with flow matching. Our approach generates LR images with realistic artifacts at unseen degradation levels, which facilitates the creation of large-scale, real-world SR training datasets. Comprehensive quantitative and qualitative assessments verify that our synthetic LR images accurately replicate real-world degradations. Furthermore, both traditional and arbitrary-scale SR models trained using our datasets consistently yield much better HR outcomes.
\end{abstract}

\begin{links}
    \link{Code}{https://github.com/present091/DegFlow}
    % \link{Datasets}{https://aaai.org/example/datasets}
    % \link{Extended version}{https://aaai.org/example/extended-version}
\end{links}

\vspace{-4mm}
\section{Introduction}

Single image super-resolution (SR) aims to reconstruct a high-resolution (HR) image from a low-resolution (LR) observation. 
Recent deep learning-based methods~\cite{vdsr, rcan, swinir, hat, mambair, mambairv2, idf} achieve strong performance in supervised settings by learning an end-to-end mapping from synthetic LR inputs to HR outputs. 
However, most SR models are trained and evaluated on LR–HR pairs generated with simple operators such as bicubic downsampling. 
As a result, they often perform poorly on real photographs, where degradations combine unknown blur, noise, and compression artifacts that are not captured by such synthetic pipelines.

One approach to reducing the distribution gap is to augment training data with handcrafted degradation pipelines that consist of blur kernels, noise, downsampling, and compression artifacts~\cite{real_esrgan, bsrgan}. 
Although these pipelines improve robustness, they still cannot represent the complex properties of real-world degradations~\cite{interflow, realdgen}.
Another line of research acquires paired HR-LR images with physical devices using DSLR cameras with zoom lenses~\cite{realsr, drealsr, coz, realarbisr}.
While such datasets provide plausible real-world degradations, the collection process is labor-intensive and limits both scale diversity and scene variety.
To alleviate these limitations, several recent promising studies learn a degradation model from a small set of real HR-LR images
and then synthesize additional LR images to boost realistic SR performance~\cite{deflow, interflow, realdgen}.

\begin{figure}[]
% \vspace*{5mm}
\begin{center}
\centerline{\includegraphics[width=1.0\columnwidth]{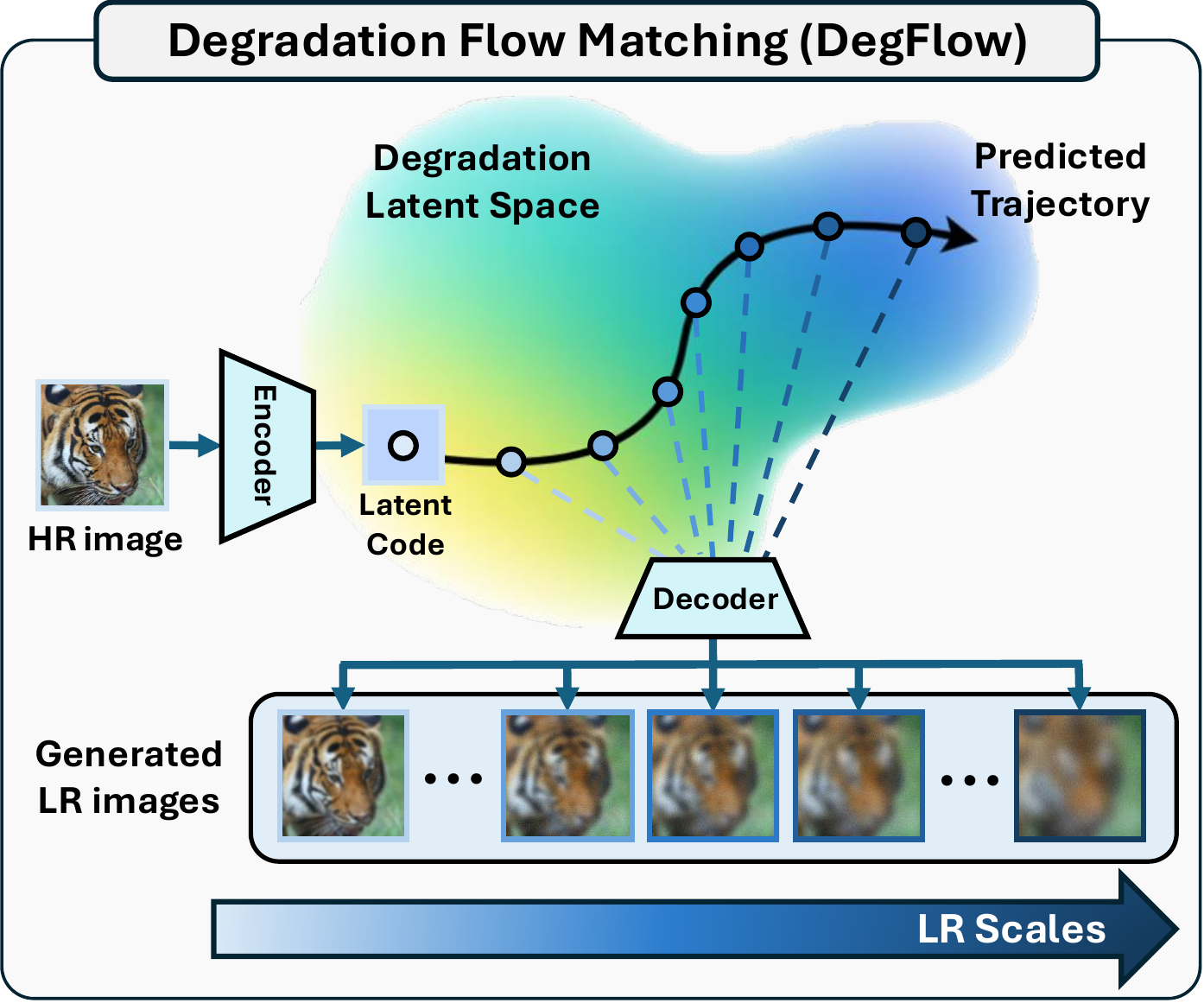}}
\caption{\framework{} generates real-world LR images across continuous scales by modeling degradation trajectories in a learned latent space. The generated LR images are used to train arbitrary SR models for high-quality restoration.}
\label{fig:teaser}
\vspace{-10mm}
\end{center}
\end{figure}

Motivated by these approaches, we introduce \textbf{\framework{}}, a novel degradation modeling framework, as shown in~\ffigref{fig:teaser}, that learns real-world degradations 
from a small set of discrete scale factors (\eg$\times2$, $\times4$) with the corresponding HR image and synthesizes LR images at \emph{unseen} continuous scales (\eg $\times2.55$, $\times3.78$) during inference.  

In \ttabref{tab:cmp_method}, we compare the proposed \framework{} with representative SR dataset generation methods.
Unlike Real-ESRGAN~\cite{real_esrgan} and BSRGAN~\cite{bsrgan}, which rely on handcrafted operators (\eg Gaussian noise, blur, bicubic down/up-sampling), 
\framework{} produces more realistic degradations by modeling degradation in real-world datasets.  
Compared with DeFlow~\cite{deflow} and RealDGen~\cite{realdgen}, \framework{} offers explicit, scale-specific control, which is essential for training arbitrary-scale SR networks.  
In contrast to InterFlow, which requires paired LR images at two distinct scales, a setting that may not be applicable in real-world scenarios,
\framework{} generates LR outputs from a single HR input at inference time.

\framework{} consists of two modules: Residual Autoencoder (RAE) and Latent Flow Matching (LFM), which are trained sequentially in a two-stage pipeline inspired by latent diffusion models~\cite{sd1, sdxl}.

Specifically, the RAE maps an input image to a compact latent code, reducing computational cost and enabling direct manipulation in latent space.  
Training on paired HR-LR images embeds degradation cues directly in the latent representation, which benefits the subsequent modeling stage.

LFM learns a continuous degradation trajectory in latent space by training a flow-matching network to a \emph{natural cubic spline} 
that interpolates the sparse degradation levels available in the training data.  
In contrast to simple piece-wise linear interpolation in the latent space, the spline model better captures nonlinear geometry in latent space while ensuring the trajectory’s first derivative remains continuous, as required by ODE.
To further improve perceptual quality, we incorporate an LPIPS loss. 
Specifically, we project a predicted latent at an intermediate scale (\eg $\times 3.34$) to its nearest available scale in the training set (\eg $\times 4.0$), allowing perceptual supervision to be applied even when direct ground truth at the exact target scale is unavailable.

Given a single HR image, the trained LFM samples latent codes at arbitrary points along the predicted latent trajectory path. Subsequently, we utilize the RAE decoder to produce LR images that demonstrate realistic degradations, including those not shown during training.
Extensive experiments show that \framework{} produces more realistic degradations than prior methods, while requiring only an HR input at test time, unlike InterFlow, which depends on paired LR examples as well as an HR image.  
The synthetic datasets generated by \framework{} allow both fixed-scale and arbitrary-scale SR networks 
to achieve state-of-the-art (SOTA) performance on numerous real-world benchmark datasets.

\begin{table}[]
\centering
\caption{Comparison between SR dataset generation methods.}
\resizebox{1.0\columnwidth}{!}{
\begin{tabular}{lccc}
\toprule[0.5pt] 
\rowcolor{HeaderBlue}
\shortstack[l]{Generation\\ Method} &
\shortstack[c]{Realistic\\ LR?} &
\shortstack[c]{Arbitrary-Scale\\ Generation?} &
\shortstack[c]{Require only HR\\ for Generation?} \\ \midrule[0.2pt]  
Real-ESRGAN, BSRGAN                                         & \xm                                                      & \xm                                                               & \cm                                                                          \\
DeFlow, RealDGen                                                    & \cm                                                      & \xm                                                               & \cm                                                                          \\
InterFlow                                                   & \cm                                                      & \cm                                                               & \xm                                                                          \\
\rowcolor{gray!10}
DegFlow (Ours)                                              & \cm                                                      & \cm                                                               & \cm                                                                          \\ \bottomrule[0.5pt]
\end{tabular}
}
\label{tab:cmp_method}
\vspace*{-5mm}
\end{table}

\section{Related Work}
\label{related_works}

\subsubsection{Image Super-Resolution.}

Single-image super-resolution (SR) remains a fundamental problem in computer vision.  
Recent deep learning approaches have achieved substantial performance gains.  
RCAN~\cite{rcan}, a CNN-based SR network, introduces multi-scale skip connections that bypass low-frequency content, allowing the network to concentrate on high-frequency detail.  
SwinIR~\cite{swinir} incorporates window-based self-attention, 
which enlarges the receptive field while reducing the quadratic complexity of standard attention.  
MambaIR~\cite{mambair} applies a selective structured state-space model to model long-range dependencies with linear computational cost.

\subsubsection{Arbitrary-Scale Super-Resolution.}

The conventional SR models can handle only a discrete set of scale factors, which limits their applicability in scenarios that require continuous zoom.  
MetaSR~\cite{metasr} is the first method to handle arbitrary continuous scales through a Meta-upscale module that predicts scale-conditioned convolution weights.  
LIIF~\cite{liif} reformulates SR as an implicit neural representation, modeling the image as a continuous function of spatial coordinates and scale.  
CiaoSR~\cite{ciaosr} enhances implicit SR by aggregating visually similar features through an attention-in-attention architecture, further enlarging the receptive field.

\subsubsection{Real-World SR Dataset.}

SR models trained on bicubic downsampled synthetic LR images perform poorly on real photographs due to the inability of these operators to capture the complexity of real-world degradations.
To reduce the discrepancy between training and real-world test scenarios,
BSRGAN~\cite{bsrgan} and Real-ESRGAN~\cite{real_esrgan} propose handcrafted degradation pipelines. 
These strategies improve robustness, but they still fall short of modeling complex real degradations.

To address this problem, several works manually capture real LR-HR pairs of images using physical equipment (\eg DSLR cameras).
RealSR~\cite{realsr} and DRealSR~\cite{drealsr} capture the dataset with multiple focal lengths and then align the images across scales.  
Recent 
RealArbiSR~\cite{realarbisr} follows a similar acquisition strategy as RealSR, but collects each LR image at its native resolution.
Although these datasets accurately reflect real degradations, their collection is costly and labor-intensive.

\subsubsection{Real-World Degradation Modeling.}
Recent works reduce data-collection costs by learning the degradation process.
DeFlow~\cite{deflow} and NAFlow~\cite{kim2024srgb} model degradations in latent space using a conditional normalizing flow, 
while RealDGen~\cite{realdgen} employs a diffusion model with contrastive disentanglement for unpaired settings.
A key limitation of these methods is the absence of explicit scale control, which is essential for training implicit SR networks.
InterFlow~\cite{interflow} addresses controllability by interpolating LR latents with a normalizing flow, 
thereby synthesizing images at unseen intermediate scales.
However, InterFlow requires paired LR observations at two distinct scales, limiting its practicality in real-world scenarios.

\begin{figure*}[!ht]
\begin{center}
\centerline{\includegraphics[width=1.0\textwidth]{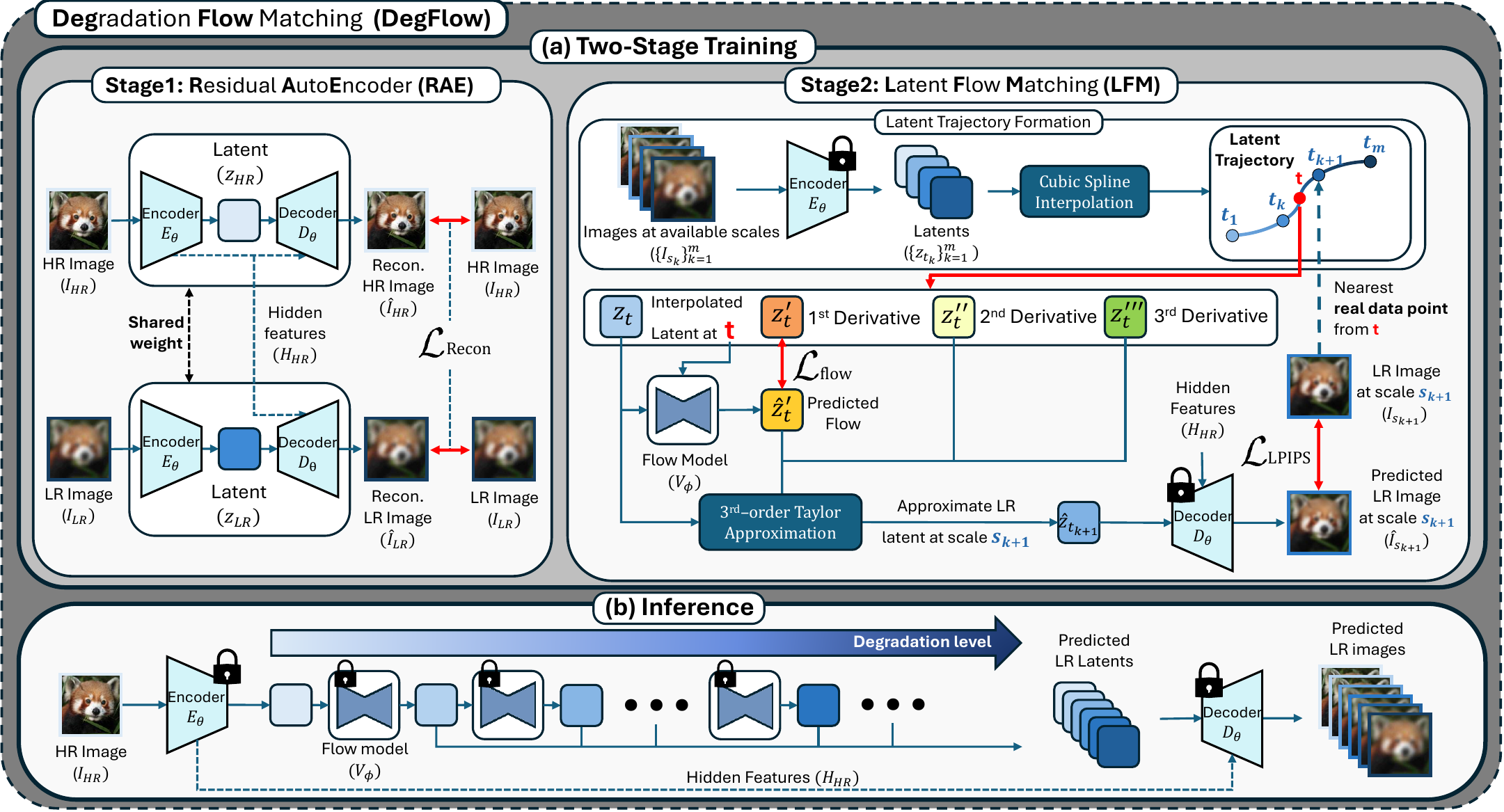}}
\caption{Overview of the proposed method. (a) Two-stage training phase. (b) Inference phase.}
\label{fig:overall_flow}
\vspace*{-10mm}
\end{center}
\end{figure*}

\section{Proposed Method}
\subsection{Preliminaries}
\label{sec:3.1}
\subsubsection{Real-World Degradation Acquisition.}

In this paper, we focus on modeling the real-world degradations that occur in real photographs acquired by DSLR cameras.
We follow prior works~\cite{realsr, drealsr, coz, realarbisr} to define the image scale $s$ by the ratio of focal lengths. We let $\mathcal{S} = \{s_k\}^{m}_{k=1}$ denote the discrete set of scales in the dataset.
We represent HR images as $I_{\mathrm{HR}}$ (or equivalently $I_{s_1}$) and their corresponding LR counterparts as $I_{\mathrm{LR}}$ (or equivalently $\{I_{s_k}\}_{k>1}$).
Lower scale values correspond to HR images that preserve finer details, while higher scale values correspond to LR images that exhibit stronger degradations, including blur, noise, and compression artifacts.

Among several real-world SR datasets, we adopt the RealSR benchmark~\cite{realsr} for training.  
RealSR provides carefully aligned LR-HR pairs at multiple scale factors. 
Specifically, each LR image is aligned to its HR counterpart by correcting misalignment caused by lens distortion and exposure variations through affine registration, followed by luminance compensation.  
The resulting image pairs 
$\{I_{\mathrm{HR}}, I_{s_2}, I_{s_3}, I_{s_4}\}$
share the same scene at the same resolution and are both geometrically and photometrically consistent, facilitating accurate modeling of real-world degradations.

Notably, we use the term \emph{degradation level} interchangeably with the scale factor $s$, following the convention in InterFlow~\cite{interflow} throughout the paper.

\subsubsection{Flow Matching.}

Flow Matching (FM)~\cite{fm, cfm, rf} is a family of generative models that learns a neural velocity field to approximate the true probability flow along a user-specified transport path.  
Below, we revisit its four essential components: ordinary differential equations (ODEs), vector fields, the FM loss, and probability paths.

First, an ODE specifies how a state $x$
% $x_t\in\mathbb{R}^d$ 
evolves with respect to time $t\in[0,1]$ as:
% https://openreview.net/pdf?id=CD9Snc73AW eq 1
\begin{equation}
dx = u(x, t) dt,
\label{eq:ode}
\end{equation}
where  
$u$
is a vector field and we will use $u(x,t)$ and $u_t(x)$ interchangeably.
Given a vector field $u$ and an initial state $x_0$, the flow (or trajectory) is given as follows:
\begin{equation} 
x_t = x_0 + \int_{0}^{t} u(x, w)dw.
\label{eq:flow}
\end{equation}
where % $u(x,t)$ 
$u(x,w)$
denotes the exact velocity that transports a target probability path  $\{p_t\}_{t\in[0,1]}$.
Pushing forward an initial probability distribution $p_0$ through $u_t$ produces $p_t$.
  
FM trains a neural field $v_{\phi}$ to regress onto $u$ so that its estimated distribution matches the target distribution $p_t$ at every timestep through the FM loss as:
% https://openreview.net/pdf?id=CD9Snc73AW eq 3
\begin{equation}
\mathcal{L}_{\text{FM}}
=\mathbb E_{t\sim\mathcal U[0,1], x\sim p_t}
\bigl\|\,v_\phi(x,t)-u(x,t)\bigr\|_2^{2},
\label{eq:fm}
\end{equation}
where time $t$ is sampled from uniform distribution ($\mathcal{U}[0,1]$).
In practice, both $u(x,t)$ and $p_t(x)$ are not given in closed form, making ~\eeqref{eq:fm} typically unsolvable. Therefore, Conditional flow matching (CFM)~\cite{cfm} is proposed to alleviate this by conditioning on an auxiliary variable $\epsilon$. Specifically, the CFM loss is given as follows:
% https://openreview.net/pdf?id=CD9Snc73AW eq 10
\begin{equation}
\small
\mathcal L_{\text{CFM}}
=\mathbb E_{t\sim\mathcal U[0,1], x\sim p_t(x|\epsilon),\epsilon\sim q(\epsilon)}
\bigl\|\,v_\phi(x,t)
-u(x,t | \epsilon)\|_2^{2},
\label{eq:cfm}
\end{equation}
where $\epsilon$ can be treated as a pair of samples (\eg HR and LR images) drawn from the joint distribution of initial (source) and target $q(\epsilon) = \pi(x_0, x_1)$. 
The conditional vector field $u(x,t | \epsilon)$ is defined according to a form of probability path $p_t(x|\epsilon)$.
A probability path is a smooth family of distributions ${p}_{t\in[0,1]}$ connecting $p_0$ and $p_1$.
Typical choices include stochastic (\eg Gaussian) bridge~\cite{fm} and deterministic (\eg Dirac) bridge~\cite{rf}.
In this work, we adopt the deterministic bridge as follows:
% https://openreview.net/pdf?id=hwnObmOTrV eq 3
\begin{equation}
p_t(x\,|\,\epsilon)=\delta\bigl(x-\mu_t(\epsilon)\bigr),
\label{eq:path_dirac}
\end{equation}
where $\delta(\cdot)$ indicates the Dirac delta function and $\mu_t(\epsilon)$ is the interpolant (\eg linear interpolation) that satisfies the boundary conditions: $\mu_0(\epsilon)=x_0$ and $\mu_1(\epsilon)=x_1$. 
As setting the intermediate variance to zero removes stochasticity (\ie $\sigma_t(\epsilon)=0$), conditional vector field $u(x,t \mid \epsilon)$ in~\eeqref{eq:cfm}  
can be derived as the first derivative of the interpolant $\mu'_t(\epsilon)$~\cite{rf}.
Consequently, the CFM loss in~\eeqref{eq:cfm} can be reformulated as
\begin{equation}
\mathcal L_{\text{CFM}}
=\mathbb E_{t\sim\mathcal U[0,1], x\sim p_t(x|\epsilon),\epsilon\sim q(\epsilon)}
\bigl\|\,v_\phi(x,t)
-\mu'_t(\epsilon)\|_2^{2},
\label{eq:re_cfm}
\end{equation}
which remains fully tractable.

\subsection{Overall Flow}
\label{sec:3.2}

We propose \framework{}, which synthesizes realistic LR images exhibiting real-world degradations while requiring only a single HR input at test time.  
As illustrated in~\ffigref{fig:overall_flow}, \framework{} comprises two components sequentially trained in a two-stage pipeline: a residual autoencoder followed by a latent flow matching model.

\subsubsection{Stage 1: Residual Autoencoder (RAE).}  
Motivated by latent diffusion studies~\cite{sd1,refusion}, 
we first train the RAE to map each image to a compact latent code with an $\mathcal{L}_2$ image reconstruction loss.  
To preserve fine details despite the high compression ratio in the latent space, we incorporate multi-scale skip connections between the encoder and decoder that propagate hidden features from the HR images. In this design, both HR and LR images are used for latent embedding, while only HR features are injected into the decoder through skip connections.

\subsubsection{Stage 2: Latent Flow Matching (LFM).}  
Once the RAE is trained and frozen, we embed paired HR-LR images in the latent space and construct trajectories that connect them.  
The FM network~\cite{song_score} learns a continuous degradation flow along each trajectory, parameterized by a time variable $t\in[0,1]$.

\subsubsection{Inference.}  
Given a single HR image, the encoder first embeds it into a latent representation.  
The FM model then evolves this latent over continuous timesteps, and the decoder transforms the evolved latents at arbitrary timesteps back into the image domain. 
By varying the timestep, \framework{} can synthesize LR outputs corresponding to intermediate scales including previously unseen degradation levels.

\subsection{Residual Autoencoder (RAE)}
\label{sec:rae}
As illustrated on the left side of \ffigref{fig:overall_flow} (a), the RAE consists of an encoder $E_{\theta}$ and a decoder $D_{\theta}$.  
Given an input image $I \in \mathbb{R}^{C \times H \times W}$, which can be an HR image $I_{\mathrm{HR}}$ or an LR image $I_{\mathrm{LR}}$, the encoder produces a compact latent representation:
$
z \;=\; E_{\theta}(I)\in\mathbb{R}^{C r^{2}\times\frac{H}{r}\times\frac{W}{r}},
$
where $C$, $H$, and $W$ indicate the channel dimensions, height, and width of the HR image, respectively, and $r$ is the spatial compression factor.
A larger $r$ reduces computational cost, but also removes high-frequency details that are essential for high-fidelity restoration~\cite{sd1, sdxl, refusion}. To mitigate this loss, we propagate multi-scale encoder features to the decoder through residual skip connections as:
\begin{equation}
\hat I \;=\; D_{\theta}\!\bigl(z;\,H_{\text{HR}}\bigr),
\end{equation}
where $H_{\text{HR}}=\{h^{(l)}_{\text{HR}}\}_{l=1}^L$ is the set of hidden features on multiple scales, and $h^{(l)}_{\text{HR}}$ denotes the hidden feature at scale level $l$ among $L$ scales.
Notably, multi-scale hidden features $H_{\text{HR}}$ are extracted from the HR image only, and thus these skip connections enable the decoder to recover fine-grained spatial details 
while leveraging the compact latent 
for efficiency.

\subsubsection{Reconstruction loss.}
Inspired by ReFusion~\cite{refusion}, the RAE is trained with a reconstruction loss $\mathcal{L}_{\text{Recon}}$ applied to HR and LR inputs.
\begin{equation}
\begin{split}
\mathcal{L}_{\text{Recon}}
=\bigl\|D_{\theta}(E_{\theta}(I_{s_1});H_{\text{HR}})-I_{s_1}\bigr\|_2^2
+\\
\bigl\|D_{\theta}(E_{\theta}(I_{s_k});H_{\text{HR}})-I_{s_k}\bigr\|_2^2,
\end{split}
\label{eq:rae_loss}
\end{equation} 
where the scale $s_k$ is drawn uniformly from the available degradation levels in the training dataset, excluding $s_1$ (\ie HR scale).
This objective ensures that the decoder can faithfully reconstruct the input images while HR features are consistently injected through skip connections, regardless of the input degradation level.
As a result, the latent space encodes only the residual information between LR and HR features (\eg degradation-specific information), providing an informative representation for the subsequent FM model.

\subsection{Latent Flow Matching (LFM)}
\label{sec:lfm}

This subsection first formalizes the probability path that links latents at different degradation levels and then introduces an auxiliary perceptual loss that further improves visual fidelity.

\subsubsection{Probability Path.}
As illustrated on the right side of~\ffigref{fig:overall_flow} (a),
each input image is first embedded into the latent space using the RAE:
\begin{equation}
z_{t_k} = E(I_{s_k}), \quad t_k = \frac{s_k - s_1}{s_m - s_1},
\end{equation}
where $t_k$ denotes the min-max normalized timestamp, and $s_1$ and $s_m$ are the minimum and maximum degradation levels in the scale set $\mathcal{S}$.
This normalization linearly maps the degradation level $s_k$ to the timestamp within the range $[0, 1]$.
For instance, consider $\mathcal{S} = \{1, 2, 4\}$, where each $s_k \in \mathcal{S}$ is distinctly matched with $t_k$, and we have $t_1=0$, $t_2=\frac{1}{3}$, and $t_4=1$.

We then construct a continuous trajectory that bridges these embedded latents across degradation levels. Several strategies can be adopted for trajectory construction. 
The simplest way to connect these marginals is a piecewise linear trajectory,
but it is suboptimal because
the latent manifold is highly nonlinear. 
Thus, linear interpolation deviates from natural-image geometry, and
the derivative $\mu'(t)$ in~\eeqref{eq:re_cfm} is discontinuous, thus violating the smoothness assumptions required by the ODE formalism~\cite{fm}. 

We therefore adopt a \textit{natural cubic spline}, which produces a trajectory with continuous first and second derivatives and a piecewise constant third derivative, whose regularity satisfies the flow matching objective.  
The deterministic mean trajectory in~\eeqref{eq:path_dirac} is thus defined as a natural cubic spline 
that interpolates over each sub-interval $[t_k,\, t_{k+1}]$ as:
\begin{equation}
\scriptsize
\mu_{t}\bigl(\epsilon\bigr)
=  a_k(\epsilon)\bigl(t-t_k\bigr)^3
 + b_k(\epsilon)\bigl(t-t_k\bigr)^2
 + c_k(\epsilon)\bigl(t-t_k\bigr)
 + d_k(\epsilon),
\end{equation}
where $t \in [t_k,\,t_{k+1}]$, $\epsilon = \{z_{t_k}\}_{k=1}^{m}$ denotes the set of latent representations at the available degradation levels,
and the coefficients $\{a_k(\epsilon), b_k(\epsilon), c_k(\epsilon), d_k(\epsilon)\}$ are obtained by solving a tridiagonal system that enforces the continuity of $\mu_{t}(\epsilon)$ and its first two derivatives, together with the natural boundary conditions $\mu_{t_1}''(\epsilon)=\mu_{t_{m}}''(\epsilon)=0$ \cite{spline}.

Given this trajectory, the LFM network predicts the velocity $\hat z'_t$, which is regressed to the ground-truth velocity using the CFM loss $\mathcal{L}_{\text{CFM}}$ in~\eeqref{eq:re_cfm}.

\subsubsection{Perceptual Loss.}

Along with the CFM loss, we introduce an additional perceptual loss to improve visual fidelity at unseen, intermediate degradation scales.
Direct supervision at these scales (\eg $\times$1.532, $\times$3.361) is infeasible
because ground-truth LR images do not exist. 
Instead, we approximate the predicted latent $\hat z_t$ at an intermediate timestep $t$ satisfying $t_k < t <t_{k+1}$ % $t\in(t_k, t_{k+1})$ 
by extrapolating toward the next degradation level $s_{k+1}$ in the set of training degradation levels $\mathcal{S} = \{s_k\}^{m}_{k=1}$. Specifically, we employ a third-order Taylor expansion for the extrapolation as follows:
\begin{equation}
\hat z_{t_{k+1}}
 = z_t
 + \hat z'_t\,\Delta t
 + \tfrac{1}{2} z''_t\,\Delta t^{2}
 + \tfrac{1}{6} z'''_t\,\Delta t^{3},
\label{eq:taylor}
\end{equation}
where $\hat z'_t$ is the predicted velocity from the LFM network, $\Delta t = t_{k+1} - t$, while $z_t$, $z''_t$, and $z'''_t$ are computed from $\mu_t(\epsilon)$.
Note that $z_t = \mu_t(\epsilon)$ due to the spline defining the deterministic mean path.
 
The extrapolated latent $\hat z_{t_{k+1}}$ corresponds to the coarser degradation level $s_{k+1}$, 
for which a ground-truth LR image $I_{s_{k+1}}$ is available. 
We can therefore compute a perceptual loss between the decoded $\hat z_{t_{k+1}}$ and $I_{s_{k+1}}$ using the LPIPS metric \cite{lpips} as:
\begin{equation}
\mathcal{L}_{\text{LPIPS}}
= \mathrm{LPIPS}\!\bigl(I_{s_{k+1}},\,D_{\theta}\!\bigl(\hat z_{t_{k+1}}\bigr)\bigr).
\end{equation}
The LPIPS loss back-propagates through~\eeqref{eq:taylor} into $\hat z'_t$ and subsequently updates the LFM network parameters $\phi$ in~\eeqref{eq:re_cfm}. This enables perceptual supervision to be applied even for intermediate scales without direct ground-truth supervision.

\subsubsection{Final Training Objective.}

The LFM network is optimized with the combined loss as:
\begin{equation}
\mathcal{L}_{\text{total}}
= \mathcal{L}_{\text{CFM}}
+ \lambda\,
  \mathcal{L}_{\text{LPIPS}},
\label{eq:fm_total_loss}
\end{equation}
where $\lambda$ balances perceptual quality and trajectory fidelity (we set $\lambda=0.1$ in all experiments).  
This joint objective encourages LFM to reproduce natural textures while faithfully modeling the underlying degradation trajectory.

\section{Experiments}
Please see the supplementary material for additional details and results. 

\subsection{Experimental Setup}
\label{sec:4.1}

\subsubsection{Implementation Details}
\hfill \break
% Please see the Supplementary Material for more details.
\noindent\textbf{RAE.}  
The RAE is trained using the Adam optimizer to minimize the reconstruction loss in Eq.~\eqref{eq:rae_loss}. Training continues for 200k iterations with a cosine-annealed learning rate schedule, decaying from $1\times10^{-4}$ to $1\times10^{-7}$. Each mini-batch contains 16 randomly cropped $256\times256$ patches with random horizontal and vertical flips for data augmentation.

\noindent\textbf{LFM.}  
The LFM network uses the Adam optimizer to minimize the CFM and LPIPS losses in Eq.~\eqref{eq:fm_total_loss} over 400k iterations. A cosine-annealed learning rate schedule decays from $2\times10^{-4}$ to $1\times10^{-7}$, with mini-batches of 32 randomly cropped $256\times256$ patches and random flips.

\subsubsection{Datasets}
\hfill \break 
\noindent\textbf{Training.}  
In all experiments, \framework{} is trained on the RealSR-V2 dataset, which contains paired images at degradation levels $\times1$, $\times2$, and $\times4$ from two DSLR camera models: Canon and Nikon.
Following InterFlow, we train on the Canon-train dataset and generate LR images from HR images of Nikon-train dataset to test the robustness of our method.

\noindent\textbf{Evaluation.}  
SR performance is evaluated on two real-world benchmarks: RealSR~\cite{realsr} and RealArbiSR~\cite{realarbisr}.  
These datasets collectively cover a wide range of camera, scene, and degradation characteristics, offering a comprehensive evaluation of generalization.

\subsection{LR Image Generation Results}
\subsubsection{Continuous Degradation Modeling Visualization.}
% \ffigref{fig:gen_lr_vis} 
We first demonstrate the capability of our~\framework{} to model continuous real-world degradations in latent space.
In \ffigref{fig:gen_lr_vis} (a), we display RealSR dataset images at discrete scales (HR, $\times$2, $\times$3, $\times$4). \ffigref{fig:gen_lr_vis} (b) shows LR images generated by \framework{} at uniformly spaced timesteps 
% $t \in [0, 1]$, 
$0\leq t \leq 1$,
using the model trained on degradation levels $\mathcal{S} = \{1, 2, 4\}$.
Our approach achieves smooth and physically consistent transitions between scales, and the synthesized images exhibit gradual variations in blur and detail loss.
These transitions closely match the characteristics of both seen levels ($\times$2, $\times$4) and unseen levels ($\times$3), indicating that \framework{} successfully learns a scale-continuous degradation manifold.

% \subsubsection{$\sbullet$ Continuous Degradation Modeling Visualization.}
\subsubsection{Timestep-Specific Degradation Analysis.}
To verify whether \framework{} accurately models degradation characteristics across the continuous trajectory, we evaluate its synthesized degradation transition at different timesteps.
\ffigref{fig:gen_lr_scores} shows the normalized PSNR, CLIP~\cite{clip}, and FID~\cite{fid} scores across different timesteps $t$ on the RealSR $\times$3 test set.
For visualization, each metric is normalized to its respective maximum value to facilitate direct comparison, and the FID scores are inverted so that higher values indicate better performance.
We observe that PSNR and FID values peak around $t \approx 0.73$ (corresponding to a degradation level of $s \approx 3.2$) and CLIP score peaks around $t \approx 0.70$ (corresponding to $s \approx 3.1$), where the synthesized degradations most closely match the real-world $\times$3 characteristics.
This result demonstrates that DegFlow successfully learns a continuous degradation manifold and captures timestep-specific degradation, both of which are essential for training arbitrary-scale SR models that require scale-specific degradations in the dataset.

\subsection{Real-world SR Performance}

\begin{figure}[!t]
\begin{center}
\centerline{\includegraphics[width=1.0\columnwidth]{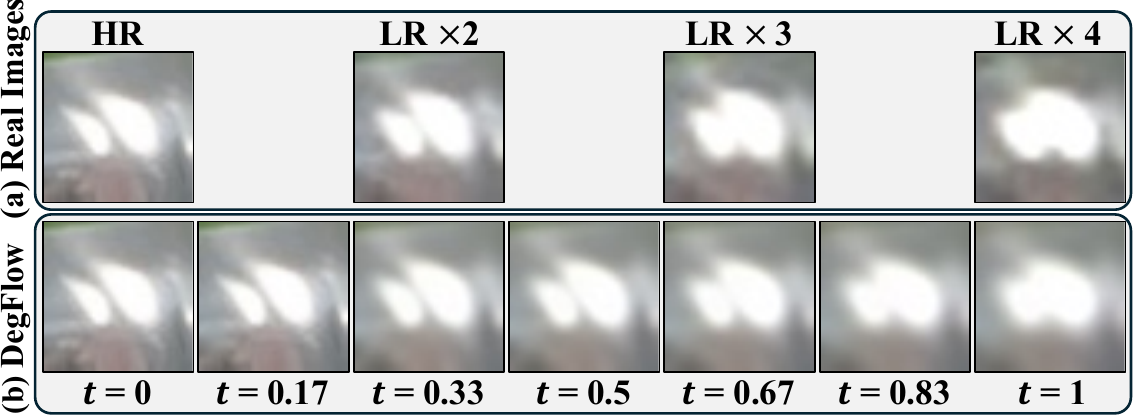}}
% \vspace{-10px}
\caption{Visualization of continuous degradation. (a) Real images from the RealSR dataset at discrete scales (HR, ×2, ×3, ×4). (b) \framework{}-generated intermediate degradations at evenly spaced timesteps $0\leq t \leq 1$.
% $t \in [0, 1]$.
}
\label{fig:gen_lr_vis}
\vspace*{-8mm}
\end{center}
\end{figure}

\begin{figure}[!t]
\begin{center}
\centerline{\includegraphics[width=1.0\columnwidth]{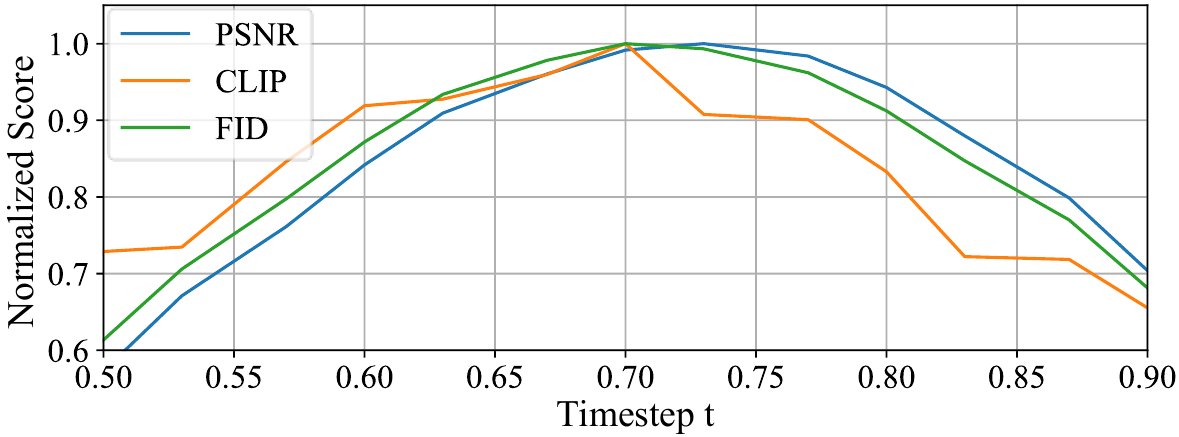}}
\vspace{-5px}
\caption{Normalized PSNR, CLIP, and FID scores across different timesteps on the RealSR $\times3$ test set.}
\label{fig:gen_lr_scores}
\vspace*{-10mm}
\end{center}
\end{figure}

\subsubsection{Fixed-Scale SR.}

\ttabref{tab:fixed_sr_result} demonstrates quantitative results on the RealSR $\times$3 test set, comparing five models: RCAN, HAN, SwinIR, HAT, and MambaIR.
First, the oracle setting is established by training on RealSR $\times$3, which directly matches the target degradation level.
Next, SR models are trained on RealSR $\times$2 and $\times$4, without using the target degradation level.
Finally, SR models undergo training using synthetic LR images generated by InterFlow and our model (Ours), ranging from $\times 2$ to $\times 4$.
Notably, InterFlow and our model are trained on only RealSR $\times$2 and $\times$4 datasets, synthesizing intermediate scales including the target scale ($\times$3).

In the results, the SR models trained with our synthesized dataset consistently outperform those trained on RealSR $\times$2, $\times$4 and InterFlow $\times$2$\sim$$\times$4, achieving higher PSNR and SSIM values while maintaining comparable or better LPIPS.
These results validate the effectiveness of our continuous degradation modeling in generating realistic and scale-continuous LR images, enabling SR networks to generalize more effectively to unseen target scales. 

In \ffigref{fig:qualitative_results}, we show SR results from HAN, HAT, and MambaIR. The results indicate that SR models trained with our synthesized LR images achieve noticeably better SR quality compared with those trained using InterFlow-generated data.

\subsubsection{Arbitrary-Scale SR.}
\ttabref{tab:arb_scale_sr} demonstrates quantitative results for arbitrary-scale SR, where we evaluate three SR models (MetaSR, LIIF, and CiaoSR).
%under different training settings.
RealSR $\times$3 refers to the ground-truth target scale dataset (oracle setting), without requiring an LR generation method.
RealSR $\times$1 represents the HR images, and is used with LR generation methods (Bicubic, BSRGAN, and Real-ESRGAN).
Bicubic denotes the conventional arbitrary-scale SR training strategy, where LR images are generated via bicubic downsampling of HR images, and Real-ESRGAN and BSRGAN synthesize LR images for training through a hand-crafted pipeline from the HR images.
InterFlow and our model (Ours) are trained on only RealSR $\times$2 and $\times$4 datasets, synthesizing intermediate-scale datasets in the range $\times$2$\sim$$\times$4 to train SR models.
Unlike InterFlow, which needs both LR and HR images, we use only HR images.

Compared with the oracle setting, our method achieves higher PSNR and lower LPIPS with comparable SSIM, indicating that it can closely approximate the upper bound without requiring ground-truth LR images at the target scale.
Moreover, our method consistently matches or surpasses InterFlow in PSNR and LPIPS across all SR models, with significant perceptual quality enhancements.
These results demonstrate that our continuous degradation modeling produces realistic and scale-consistent LR images, enabling arbitrary-scale SR networks to generalize more effectively to unseen target scales.

As shown in \ffigref{fig:qualitative_results}, the arbitrary-scale SR results (MetaSR, LIIF, CiaoSR) show that models trained with our synthesized LR images achieve clearly improved SR quality compared with those trained using InterFlow-generated data.

\begin{table}[]
\centering
\newcommand{\red}[1]{\textcolor{red}{\textbf{#1}}}
\newcommand{\blue}[1]{\textcolor{blue}{#1}}
\caption{Fixed-scale SR results on RealSR $\times 3$ test set. Best and second-best are highlighted in \red{red} and \blue{blue}.}
\vspace*{-1mm}
\resizebox{0.85\columnwidth}{!}{
\begin{tabular}{l!{\vrule width 0.4pt}l!{\vrule width 0.4pt}ccc}
 \toprule
 \rowcolor{HeaderBlue} 
 \multirow{3}{*}{Model}   & \multirow{3}{*}{Train Set} & \multicolumn{3}{c}{RealSR Test Set}    \\ 
 \rowcolor{HeaderBlue}
 \multirow{-1}{*}{Model}  & \multirow{-1}{*}{SR Train Set} &               & $\times$3                &  \\ 
 \rowcolor{HeaderBlue}  
                         &                            & PSNR$\uparrow$          & SSIM$\uparrow$           & LPIPS$\downarrow$          \\ 
 \midrule

 \multirow{4}{*}{RCAN}    
                         & RealSR $\times 3$                     & 30.68          & 0.8641          & 0.3243          \\ \cdashline{2-5}\noalign{\vskip 0.5ex} 
                         & RealSR $\times 2, \times 4$           & 30.30          & 0.8596          & 0.3281          \\
                         & InterFlow $\times2$\ssim$\times4$   & \blue{30.57}   & \blue{0.8631}   & \red{0.3155}    \\
                         & \cellcolor{gray!10}Ours $\times2$\ssim$\times4$
                         & \cellcolor{gray!10}\red{30.72}
                         & \cellcolor{gray!10}\red{0.8650}
                         & \cellcolor{gray!10}\blue{0.3221}      \\ 
 \midrule 

 \multirow{4}{*}{HAN}    
                         & RealSR $\times 3$                     & 30.76          & 0.8659          & 0.3216          \\ \cdashline{2-5}\noalign{\vskip 0.5ex} 
                         & RealSR $\times 2, \times 4$           & 30.43          & 0.8616          & 0.3261          \\
                         & InterFlow $\times 2$\ssim$\times 4$      & \blue{30.68}   & \blue{0.8644}   & \red{0.3167}    \\
                         & \cellcolor{gray!10}Ours $\times2$\ssim$\times4$
                         & \cellcolor{gray!10}\red{30.82}
                         & \cellcolor{gray!10}\red{0.8660}
                         & \cellcolor{gray!10}\blue{0.3212}      \\ 
 \midrule

 \multirow{4}{*}{SwinIR} 
                         & RealSR $\times 3$                     & 30.69          & 0.8647          & 0.3217          \\ \cdashline{2-5}\noalign{\vskip 0.5ex} 
                         & RealSR $\times 2, \times 4$           & 30.23          & 0.8597          & 0.3255          \\
                         & InterFlow $\times2$\ssim$\times4$   & \blue{30.56}   & \blue{0.8634}   & \red{0.3166}    \\
                         & \cellcolor{gray!10}Ours $\times2$\ssim$\times4$
                         & \cellcolor{gray!10}\red{30.78}
                         & \cellcolor{gray!10}\red{0.8658}
                         & \cellcolor{gray!10}\blue{0.3193}      \\ 
 \midrule

 \multirow{4}{*}{HAT}    
                         & RealSR $\times 3$                     & 30.71          & 0.8645          & 0.3221          \\ \cdashline{2-5}\noalign{\vskip 0.5ex}  
                         & RealSR $\times 2, \times 4$           & 30.39          & 0.8607          & 0.3248          \\
                         & InterFlow $\times2$\ssim$\times4$   & \blue{30.65}   & \blue{0.8645}   & \red{0.3135}    \\
                         & \cellcolor{gray!10}Ours $\times2$\ssim$\times4$
                         & \cellcolor{gray!10}\red{30.86}
                         & \cellcolor{gray!10}\red{0.8668}
                         & \cellcolor{gray!10}\blue{0.3186}      \\ 
 \midrule

 \multirow{4}{*}{MambaIR}
                         & RealSR $\times 3$                     & 30.62          & 0.8636          & 0.3208          \\ \cdashline{2-5}\noalign{\vskip 0.5ex}  
                         & RealSR $\times 2, \times 4$           & 30.29          & 0.8660          & 0.3240          \\
                         & InterFlow $\times2$\ssim$\times4$   & \blue{30.51}   & \blue{0.8625}   & \red{0.3138}    \\
                         & \cellcolor{gray!10}Ours $\times2$\ssim$\times4$
                         & \cellcolor{gray!10}\red{30.73}
                         & \cellcolor{gray!10}\red{0.8686}
                         & \cellcolor{gray!10}\blue{0.3152}      \\ 
 \bottomrule
\end{tabular}
}
\label{tab:fixed_sr_result}
\vspace*{-5mm}
\end{table}

\begin{table*}[]
\centering
\newcommand{\red}[1]{\textcolor{red}{\textbf{#1}}}
\newcommand{\blue}[1]{\textcolor{blue}{#1}}
\newcommand{\lpv}[1]{\multicolumn{1}{c!{\vrule width 0.4pt}}{#1}}
\caption{Arbitrary-scale SR results on  RealSR $\times 3$ test set. Best and second-best are highlighted in \noindent\red{red} and \blue{blue}.} 
\vspace{-1mm}
\resizebox{0.9\textwidth}{!}{
\begin{tabular}{l!{\vrule width 0.4pt}l!{\vrule width 0.4pt}l!{\vrule width 0.4pt}ccc!{\vrule width 0.4pt}ccccccccc}
\toprule
\rowcolor{HeaderBlue}
\multirow{3}{*}{Model}   & \multirow{3}{*}{Train set} & \multirow{3}{*}{LR Generation Method} & \multicolumn{3}{c!{\vrule width 0.4pt}}{RealSR Test Set} & \multicolumn{9}{c}{RealArbiSR Test Set}                                       \\ \rowcolor{HeaderBlue} 
%\cline{4-15} 
\multirow{-1}{*}{Model}  & \multirow{-1}{*}{Generation Train set} & \multirow{-1}{*}{LR Generation Method} & \multicolumn{3}{c!{\vrule width 0.4pt}}{$\times$3}              & \multicolumn{3}{c!{\vrule width 0.4pt}}{$\times$2.5} & \multicolumn{3}{c!{\vrule width 0.4pt}}{$\times$3}  & \multicolumn{3}{c}{$\times$3.5} \\ \rowcolor{HeaderBlue}
                         &                                       &                                       & PSNR$\uparrow$ & SSIM$\uparrow$ & LPIPS$\downarrow$ & PSNR$\uparrow$ & SSIM$\uparrow$ & \multicolumn{1}{c!{\vrule width 0.4pt}}{LPIPS$\downarrow$} & PSNR$\uparrow$ & SSIM$\uparrow$ & \multicolumn{1}{c!{\vrule width 0.4pt}}{LPIPS$\downarrow$} & PSNR$\uparrow$ & SSIM$\uparrow$ & LPIPS$\downarrow$ \\  
\midrule   

\multirow{6}{*}{MetaSR} 
                         & RealSR $\times3$                  & None (Oracle)                         & 30.43      & 0.8572     & 0.3311     & 29.65  & 0.8679 & \lpv{0.3330} & 29.58 & 0.8338 & \lpv{0.3557} & 28.15  & 0.7974 & 0.3996 \\ \cdashline{2-15}\noalign{\vskip 0.5ex}
                         & RealSR $\times1$                  & Bicubic (Baseline)                    & 28.99      & 0.8165     & 0.3488     & 30.05  & 0.8473 & \lpv{0.3087} & 28.77 & 0.8042 & \lpv{0.3544} & 27.86  & 0.7711 & 0.3886 \\
                         & RealSR $\times1$                  & BSRGAN                                & 28.15      & 0.8114     & 0.3867     & 28.41  & 0.8326 & \lpv{0.3679} & 27.41 & 0.7932 & \lpv{0.3971} & 26.76  & 0.7625 & 0.4199 \\
                         & RealSR $\times1$                  & Real-ESRGAN                           & 26.90      & 0.8077     & 0.3813     & 27.32  & 0.8177 & \lpv{0.3738} & 26.50 & 0.7821 & \lpv{0.4014} & 25.94  & 0.7305 & 0.4380 \\
                         & RealSR $\times2, \times4$         & InterFlow                             & \blue{30.42}      & \red{0.8569}     & \blue{0.3222}     & \blue{30.71}  & \blue{0.8703} & \lpv{\blue{0.3099}} & \blue{29.40} & \blue{0.8302} & \lpv{\blue{0.3504}} & \blue{28.50}  & \blue{0.7983} & \blue{0.3828} \\
                         & RealSR $\times2, \times4$         & \cellcolor{gray!10}Ours               & \cellcolor{gray!10}\red{30.58}      & \cellcolor{gray!10}\blue{0.8565}     & \lpv{\cellcolor{gray!10}\red{0.3190}}     & \cellcolor{gray!10}\red{30.88}  & \cellcolor{gray!10}\red{0.8713} & \lpv{\cellcolor{gray!10}\red{0.2995}} & \cellcolor{gray!10}\red{29.63} & \cellcolor{gray!10}\red{0.8321} & \lpv{\cellcolor{gray!10}\red{0.3429}} & \cellcolor{gray!10}\red{28.71}  & \cellcolor{gray!10}\red{0.8008} & \cellcolor{gray!10}\red{0.3780} \\  
\midrule

\multirow{6}{*}{LIIF}    
                         & RealSR $\times3$                  & None (Oracle)                         & 30.43      & 0.8578     & 0.3324     & 30.71  & 0.8718 & \lpv{0.3222} & 29.56 & 0.8336 & \lpv{0.3579} & 28.66  & 0.8028 & 0.3861 \\ \cdashline{2-15}\noalign{\vskip 0.5ex}
                         & RealSR $\times1$                  & Bicubic (Baseline)                    & 29.00      & 0.8167     & 0.3290     & 30.04  & 0.8472 & \lpv{0.3096} & 28.76 & 0.8042 & \lpv{0.3550} & 27.86  & 0.7711 & 0.3892 \\
                         & RealSR $\times1$                  & BSRGAN                                & 28.23      & 0.8133     & 0.3875     & 28.28  & 0.8303 & \lpv{0.3719} & 27.32 & 0.7912 & \lpv{0.4009} & 26.75  & 0.7617 & 0.4242 \\
                         & RealSR $\times1$                  & Real-ESRGAN                           & 27.07      & 0.8090     & 0.3817     & 27.24  & 0.8135 & \lpv{0.3755} & 26.36 & 0.7777 & \lpv{0.4025} & 25.73  & 0.7492 & 0.4243 \\
                         & RealSR $\times2, \times4$         & InterFlow                             & \blue{30.44}      & \red{0.8581}     & \blue{0.3263}     & \blue{30.70}  & \blue{0.8705} & \lpv{\blue{0.3144}} & \blue{29.38} & \blue{0.8307} & \lpv{\blue{0.3547}} & \blue{28.44}  & \blue{0.7985} & \blue{0.3860} \\
                         & RealSR $\times2, \times4$         & \cellcolor{gray!10}Ours               & \cellcolor{gray!10}\red{30.61}      & \cellcolor{gray!10}\blue{0.8577}     & \cellcolor{gray!10}\red{0.3251}     & \cellcolor{gray!10}\red{30.99}  & \cellcolor{gray!10}\red{0.8729} & \lpv{\cellcolor{gray!10}\red{0.3105}} & \cellcolor{gray!10}\red{29.74} & \cellcolor{gray!10}\red{0.8341} & \lpv{\cellcolor{gray!10}\red{0.3517}} & \cellcolor{gray!10}\red{28.78}  & \cellcolor{gray!10}\red{0.8027} & \cellcolor{gray!10}\red{0.3845} \\  
\midrule

\multirow{6}{*}{CiaoSR}  
                         & RealSR $\times3$                  & None (Oracle)                         & 30.65      & 0.8609     & 0.3251     & 30.61  & 0.8705 & \lpv{0.3105} & 29.59 & 0.8339 & \lpv{0.3487} & 28.54  & 0.8011 & 0.3810 \\ \cdashline{2-15}\noalign{\vskip 0.5ex}
                         & RealSR $\times1$                  & Bicubic (Baseline)                    & 28.98      & 0.8160     & 0.3496     & 30.04  & 0.8472 & \lpv{0.3079} & 28.76 & 0.8037 & \lpv{0.3545} & 27.85  & 0.7708 & 0.3881 \\
                         & RealSR $\times1$                  & BSRGAN                                & 28.55      & 0.8288     & 0.3638     & 28.88  & 0.8443 & \lpv{0.3490} & 27.90 & 0.8046 & \lpv{0.3797} & 27.29  & 0.7755 & 0.4069 \\
                         & RealSR $\times1$                  & Real-ESRGAN                           & 27.48      & 0.8200     & 0.3696     & 27.86  & 0.8341 & \lpv{0.3539} & 26.95 & 0.7953 & \lpv{0.3823} & 26.31  & 0.7664 & 0.4091 \\
                         & RealSR $\times2, \times4$         & InterFlow                             & \blue{30.52}      & \blue{0.8590}     & \blue{0.3162}     & \blue{30.69}  & \blue{0.8702} & \lpv{\red{0.3057}} & \blue{29.36} & \blue{0.8298} & \lpv{\red{0.3408}} & \blue{28.52}  & \blue{0.8000} & \blue{0.3795} \\
                         & RealSR $\times2, \times4$         & \cellcolor{gray!10}Ours               & \cellcolor{gray!10}\red{30.70}      & \cellcolor{gray!10}\red{0.8590}     & \cellcolor{gray!10}\red{0.3153}     & \cellcolor{gray!10}\red{31.03}  & \cellcolor{gray!10}\red{0.8739} & \lpv{\cellcolor{gray!10}\blue{0.3059}} & \cellcolor{gray!10}\red{29.58} & \cellcolor{gray!10}\red{0.8318} & \lpv{\cellcolor{gray!10}\blue{0.3439}} & \cellcolor{gray!10}\red{28.77}  & \cellcolor{gray!10}\red{0.8032} & \cellcolor{gray!10}\red{0.3793} \\ 
\bottomrule
\end{tabular}
}
\label{tab:arb_scale_sr}
\end{table*}

\begin{figure*}[!ht]
\begin{center}
\vspace{-5px}
\centerline{\includegraphics[width=0.9\textwidth]{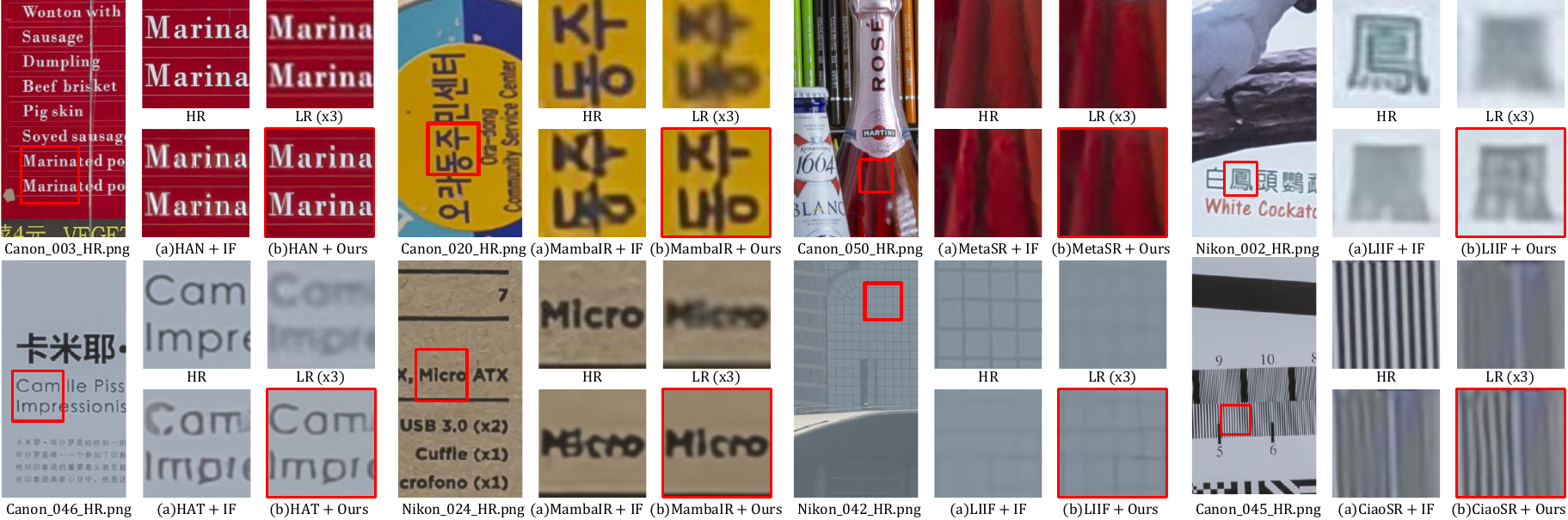}}
% \vspace{-10px}
\caption{ 
Qualitative comparisons on the RealSR $\times3$ dataset. Fixed-scale SR results (HAN, HAT, MambaIR) and arbitrary-scale SR results (MetaSR, LIIF, CiaoSR) trained with either InterFlow (IF) generated LR (a) or our synthesized LR (b) are compared.
 }
\label{fig:qualitative_results}
\vspace*{-10mm}
\end{center}
\end{figure*}

\begin{table}[]
\centering
\newcommand{\red}[1]{\textcolor{red}{\textbf{#1}}}
\caption{Impact of each component on RealSR $\times$3 test set. 
%Starting from the baseline, we incrementally incorporate our proposed components.
}
\vspace{-2mm}
\resizebox{1.0\columnwidth}{!}{
\begin{tabular}{l!{\vrule width 0.4pt}ccc}
\toprule
\rowcolor{HeaderBlue}
\multirow{2}{*}{}                        & \multicolumn{3}{c}{RealSR $\times 3$ Test Set} \\ %\cmidrule{2-4}  
\rowcolor{HeaderBlue}
\multirow{-2}{*}{Methods}                       & PSNR↑      & SSIM↑       & LPIPS↓      \\  \midrule
Baseline (Piecewise Linear Trajectory)          & 30.58      & 0.8640      & 0.3214      \\
(+) Nonlinear Trajectory (Natural Cubic Spline) & 30.68      & 0.8652      & 0.3209      \\
(+) LPIPS 3rd-order Taylor Approx.              & 30.81      & 0.8662      & 0.3200      \\
\rowcolor{gray!10}(+) RAE's HR Features Skip Connection           & \red{30.86}      & \red{0.8668}      & \red{0.3186}      \\  \bottomrule
\end{tabular}
\label{tab:abl_component}
}
\vspace*{-5mm}
\end{table}

\subsection{Ablation Study}
For the ablation study, we measure the performance of the HAT SR network on the RealSR $\times$3 test set.

\subsubsection{Effect of Proposed Components.}
In~\ttabref{tab:abl_component}, we show the effectiveness of each component in our framework.
We begin with a baseline variant of~\framework{} that adopts a piecewise linear trajectory to model the latent degradation path. 
Next, using the natural cubic spline for a nonlinear trajectory model improves PSNR and LPIPS, 
demonstrating the benefits of smooth path modeling for real-world degradations.
Then, incorporating the 3rd-order Taylor approximation for the LPIPS-based supervision further improves perceptual fidelity.
In addition, introducing skip connections from high-resolution features in the RAE further enhances the reconstruction quality, showing the best performance across all metrics. 
These results validate the importance of each proposed component 
and highlight their complementary contributions to both perceptual and distortion-based performance.

\subsubsection{Effect of External HR Dataset.}
We further investigate whether synthesizing the SR training set with external HR images can enhance SR performance. In particular, we leverage the DIV2K~\cite{div2k} dataset, which contains high-quality HR images, to generate synthetic LR images for scales in the range $\times 2$ to $\times 4$ using our framework. Notably, our approach can synthesize these intermediate degradations directly from HR images, in contrast to InterFlow, which requires paired LR images captured at multiple degradation levels. As shown in~\ttabref{tab:abl_add_hr_dataset}, training the HAT network with the augmented dataset consistently improves PSNR, SSIM, and perceptual quality on the RealSR $\times 3$ test set. Generating high-quality degradations from only HR data offers a practical benefit when paired LR data is unavailable.

\begin{table}[!t]
\centering
\newcommand{\red}[1]{\textcolor{red}{\textbf{#1}}}
\caption{Impact of using external HR images.}
\vspace{-2mm}
\resizebox{0.95\columnwidth}{!}{
\begin{tabular}{l!{\vrule width 0.4pt}ccc}
\toprule
\rowcolor{HeaderBlue}
\multirow{2}{*}{}                     & \multicolumn{3}{c}{RealSR $\times 3$ Test Set} \\ %\cmidrule{2-4} 
\rowcolor{HeaderBlue}
\multirow{-2}{*}{Train Set}           & PSNR↑      & SSIM↑       & LPIPS↓      \\ \midrule
DegFlow $\times2$\ssim$\times4$              & 30.86      & 0.8668      & 0.3186      \\
\rowcolor{gray!10}(+) Additional Synthetic Dataset (DIV2K) & \red{31.00}      & \red{0.8673}      & \red{0.3180}      \\ \bottomrule
\end{tabular}
}
\label{tab:abl_add_hr_dataset}
\vspace*{-5mm}
\end{table}

\vspace*{-2mm}
\section{Conclusion}
We introduce DegFlow, a novel continuous degradation modeling framework for real-world super-resolution. 
Unlike previous methods that rely on handcrafted degradation pipelines or require paired low-resolution inputs for generation, DegFlow learns a degradation manifold in latent space from only discrete real-world HR-LR pairs and synthesizes realistic degradations at arbitrary, unseen scales using only high-resolution images. 
By combining a residual autoencoder with latent flow matching, DegFlow effectively captures the nonlinear geometry of real-world degradations while maintaining explicit degradation level control.
Experiments show that SR networks trained on our synthetic datasets consistently outperform those trained with existing generation methods in both fidelity and perceptual quality.

\section{Acknowledgments}
This work was supported by National Research Foundation of Korea (NRF) grant funded by the Korea government (MSIT) (RS-2023-00222776), and Institute of Information communications Technology Planning Evaluation (IITP) grant funded by the Korea government (MSIT) (No.2022- 0-00156, Fundamental research on continual
meta-learning for quality enhancement of casual videos and their 3D metaverse transformation) and Institute of Information \& communications Technology Planning \& Evaluation (IITP) grant funded by the Korea government(MSIT) (No.RS-2020-II201373, Artificial Intelligence Graduate School Program(Hanyang University)) and the research fund of Hanyang University(HY-2025).

\bibliography{aaai2026}

@String(CVPR= {IEEE Conf. Comput. Vis. Pattern Recog.})

@String(ICCV= {Int. Conf. Comput. Vis.})

@String(ECCV= {Eur. Conf. Comput. Vis.})

@String(NeurIPS= {Adv. Neural Inform. Process. Syst.})

@String(ICLR = {Int. Conf. Learn. Represent.})

@String(CVPRW= {IEEE Conf. Comput. Vis. Pattern Recog. Worksh.})

@String(CVPR  = {CVPR})

@String(ICCV  = {ICCV})

@String(ECCV  = {ECCV})

@String(NeurIPS  = {NeurIPS})

@String(ICLR  = {ICLR})

@String(CVPRW= {CVPRW})

@inproceedings{realsr,
  title={Toward real-world single image super-resolution: A new benchmark and a new model},
  author={Cai, Jianrui and Zeng, Hui and Yong, Hongwei and Cao, Zisheng and Zhang, Lei},
  booktitle={ICCV},
  year={2019}
}

@inproceedings{drealsr,
  title={Component divide-and-conquer for real-world image super-resolution},
  author={Wei, Pengxu and Xie, Ziwei and Lu, Hannan and Zhan, Zongyuan and Ye, Qixiang and Zuo, Wangmeng and Lin, Liang},
  booktitle={ECCV},
  year={2020},
}

@inproceedings{coz,
  title={Continuous optical zooming: A benchmark for arbitrary-scale image super-resolution in real world},
  author={Fu, Huiyuan and Peng, Fei and Li, Xianwei and Li, Yejun and Wang, Xin and Ma, Huadong},
  booktitle={CVPR},
  year={2024}
}

@inproceedings{realarbisr,
  title={Learning Dual-Level Deformable Implicit Representation for Real-World Scale Arbitrary Super-Resolution},
  author={Li, Zhiheng and Li, Muheng and Fan, Jixuan and Chen, Lei and Tang, Yansong and Lu, Jiwen and Zhou, Jie},
  booktitle={ECCV},
  year={2024},
}

@inproceedings{vdsr,
  title={Accurate image super-resolution using very deep convolutional networks},
  author={Kim, Jiwon and Lee, Jung Kwon and Lee, Kyoung Mu},
  booktitle={CVPR},
  year={2016}
}

@inproceedings{rcan,
  title={Image super-resolution using very deep residual channel attention networks},
  author={Zhang, Yulun and Li, Kunpeng and Li, Kai and Wang, Lichen and Zhong, Bineng and Fu, Yun},
  booktitle={ECCV},
  year={2018}
}

@inproceedings{swinir,
  title={Swinir: Image restoration using swin transformer},
  author={Liang, Jingyun and Cao, Jiezhang and Sun, Guolei and Zhang, Kai and Van Gool, Luc and Timofte, Radu},
  booktitle={ICCV},
  year={2021}
}

@inproceedings{hat,
  title={Activating more pixels in image super-resolution transformer},
  author={Chen, Xiangyu and Wang, Xintao and Zhou, Jiantao and Qiao, Yu and Dong, Chao},
  booktitle={CVPR},
  year={2023}
}

@inproceedings{mambair,
  title={MambaIR: A simple baseline for image restoration with state-space model},
  author={Guo, Hang and Li, Jinmin and Dai, Tao and Ouyang, Zhihao and Ren, Xudong and Xia, Shu-Tao},
  booktitle={ECCV},
  year={2024}
}

@inproceedings{mambairv2,
  title={MambaIRv2: Attentive state space restoration},
  author={Guo, Hang and Guo, Yong and Zha, Yaohua and Zhang, Yulun and Li, Wenbo and Dai, Tao and Xia, Shu-Tao and Li, Yawei},
  booktitle={CVPR},
  year={2025}
}

@inproceedings{han,
  title={Single image super-resolution via a holistic attention network},
  author={Niu, Ben and Wen, Weilei and Ren, Wenqi and Zhang, Xiangde and Yang, Lianping and Wang, Shuzhen and Zhang, Kaihao and Cao, Xiaochun and Shen, Haifeng},
  booktitle={ECCV},
  year={2020}
}

@inproceedings{metasr,
  title={Meta-SR: A magnification-arbitrary network for super-resolution},
  author={Hu, Xuecai and Mu, Haoyuan and Zhang, Xiangyu and Wang, Zilei and Tan, Tieniu and Sun, Jian},
  booktitle={CVPR},
  year={2019}
}

@inproceedings{liif,
  title={Learning continuous image representation with local implicit image function},
  author={Chen, Yinbo and Liu, Sifei and Wang, Xiaolong},
  booktitle={CVPR},
  year={2021}
}

@inproceedings{ciaosr,
  title={Ciaosr: Continuous implicit attention-in-attention network for arbitrary-scale image super-resolution},
  author={Cao, Jiezhang and Wang, Qin and Xian, Yongqin and Li, Yawei and Ni, Bingbing and Pi, Zhiming and Zhang, Kai and Zhang, Yulun and Timofte, Radu and Van Gool, Luc},
  booktitle={CVPR},
  year={2023}
}

@inproceedings{real_esrgan,
  title={Real-ESRGAN: Training real-world blind super-resolution with pure synthetic data},
  author={Wang, Xintao and Xie, Liangbin and Dong, Chao and Shan, Ying},
  booktitle={ICCV},
  year={2021}
}

@inproceedings{bsrgan,
  title={Designing a practical degradation model for deep blind image super-resolution},
  author={Zhang, Kai and Liang, Jingyun and Van Gool, Luc and Timofte, Radu},
  booktitle={ICCV},
  year={2021}
}

@inproceedings{deflow,
  title={Deflow: Learning complex image degradations from unpaired data with conditional flows},
  author={Wolf, Valentin and Lugmayr, Andreas and Danelljan, Martin and Van Gool, Luc and Timofte, Radu},
  booktitle={CVPR},
  year={2021}
}

@inproceedings{interflow,
  title={Learning controllable degradation for real-world super-resolution via constrained flows},
  author={Park, Seobin and Kim, Dongjin and Baik, Sungyong and Kim, Tae Hyun},
  booktitle={ICML},
  year={2023}
}

@inproceedings{realdgen,
  title={Towards realistic data generation for real-world super-resolution},
  author={Peng, Long and Li, Wenbo and Pei, Renjing and Ren, Jingjing and Xu, Jiaqi and Wang, Yang and Cao, Yang and Zha, Zheng-Jun},
  booktitle={ICLR},
  year={2025}
}

@inproceedings{fm,
  title={Flow matching for generative modeling},
  author={Lipman, Yaron and Chen, Ricky TQ and Ben-Hamu, Heli and Nickel, Maximilian and Le, Matt},
  booktitle={ICLR},
  year={2023}
}

@inproceedings{cfm,
  title={Improving and generalizing flow-based generative models with minibatch optimal transport},
  author={Tong, Alexander and FATRAS, Kilian and Malkin, Nikolay and Huguet, Guillaume and Zhang, Yanlei and Rector-Brooks, Jarrid and Wolf, Guy and Bengio, Yoshua},
  booktitle={TMLR},
  year={2024}
}

@inproceedings{rf,
  title={Flow straight and fast: Learning to generate and transfer data with rectified flow},
  author={Liu, Xingchao and Gong, Chengyue and Liu, Qiang},
  booktitle={ICLR},
  year={2023}
}

@book{spline,
  title={A practical guide to splines},
  author={De Boor, Carl},
  volume={27},
  year={1978},
  publisher={Springer}
}

@inproceedings{sd1,
  title={High-resolution image synthesis with latent diffusion models},
  author={Rombach, Robin and Blattmann, Andreas and Lorenz, Dominik and Esser, Patrick and Ommer, Bj{\"o}rn},
  booktitle={CVPR},
  year={2022}
}

@inproceedings{sdxl,
  title={Sdxl: Improving latent diffusion models for high-resolution image synthesis},
  author={Podell, Dustin and English, Zion and Lacey, Kyle and Blattmann, Andreas and Dockhorn, Tim and M{\"u}ller, Jonas and Penna, Joe and Rombach, Robin},
  booktitle={ICML},
  year={2024}
}

@inproceedings{refusion,
  title={Refusion: Enabling large-size realistic image restoration with latent-space diffusion models},
  author={Luo, Ziwei and Gustafsson, Fredrik K and Zhao, Zheng and Sj{\"o}lund, Jens and Sch{\"o}n, Thomas B},
  booktitle={CVPR},
  year={2023}
}

@inproceedings{song_score,
  title={Score-based generative modeling through stochastic differential equations},
  author={Song, Yang and Sohl-Dickstein, Jascha and Kingma, Diederik P and Kumar, Abhishek and Ermon, Stefano and Poole, Ben},
  booktitle={ICLR},
  year={2021}
}

@article{ssim,
  title={Image quality assessment: from error visibility to structural similarity},
  author={Wang, Zhou and Bovik, Alan C and Sheikh, Hamid R and Simoncelli, Eero P},
  journal={IEEE Trans. Image Process.},
  volume={13},
  number={4},
  pages={600--612},
  year={2004},
  publisher={IEEE}
}

@inproceedings{lpips,
  title={The unreasonable effectiveness of deep features as a perceptual metric},
  author={Zhang, Richard and Isola, Phillip and Efros, Alexei A and Shechtman, Eli and Wang, Oliver},
  booktitle={CVPR},
  year={2018}
}

@inproceedings{div2k,
  title={NTIRE 2017 challenge on single image super-resolution: Dataset and study},
  author={Agustsson, Eirikur and Timofte, Radu},
  booktitle={CVPRW},
  year={2017}
}

@inproceedings{clip,
  title={Learning transferable visual models from natural language supervision},
  author={Radford, Alec and Kim, Jong Wook and Hallacy, Chris and Ramesh, Aditya and Goh, Gabriel and Agarwal, Sandhini and Sastry, Girish and Askell, Amanda and Mishkin, Pamela and Clark, Jack and others},
  booktitle={ICML},
  year={2021}
}

@inproceedings{fid,
  title={GANs trained by a two time-scale update rule converge to a local nash equilibrium},
  author={Heusel, Martin and Ramsauer, Hubert and Unterthiner, Thomas and Nessler, Bernhard and Hochreiter, Sepp},
  booktitle={NeurIPS},
  year={2017}
}

@article{rk45_1,
  title={A family of embedded Runge-Kutta formulae},
  author={Dormand, John R and Prince, Peter J},
  journal={J. Comput. Appl. Math.},
  volume={6},
  number={1},
  pages={19--26},
  year={1980},
  publisher={Elsevier}
}

@article{rk45_2,
  title={Some practical runge-kutta formulas},
  author={Shampine, Lawrence F},
  journal={Math. Comput.},
  volume={46},
  number={173},
  pages={135--150},
  year={1986}
}

@inproceedings{nafnet,
  title={Simple baselines for image restoration},
  author={Chen, Liangyu and Chu, Xiaojie and Zhang, Xiangyu and Sun, Jian},
  booktitle={ECCV},
  year={2022}
}

@article{layernorm,
  title={Layer normalization},
  author={Ba, Jimmy Lei and Kiros, Jamie Ryan and Hinton, Geoffrey E},
  journal={arXiv preprint arXiv:1607.06450},
  year={2016}
}

@article{gelu,
  title={Gaussian error linear units (gelus)},
  author={Hendrycks, Dan and Gimpel, Kevin},
  journal={arXiv preprint arXiv:1606.08415},
  year={2016}
}

@ARTICLE{scipy,
  author  = {Virtanen, Pauli and Gommers, Ralf and Oliphant, Travis E. and
            Haberland, Matt and Reddy, Tyler and Cournapeau, David and
            Burovski, Evgeni and Peterson, Pearu and Weckesser, Warren and
            Bright, Jonathan and {van der Walt}, St{\'e}fan J. and
            Brett, Matthew and Wilson, Joshua and Millman, K. Jarrod and
            Mayorov, Nikolay and Nelson, Andrew R. J. and Jones, Eric and
            Kern, Robert and Larson, Eric and Carey, C J and
            Polat, {\.I}lhan and Feng, Yu and Moore, Eric W. and
            {VanderPlas}, Jake and Laxalde, Denis and Perktold, Josef and
            Cimrman, Robert and Henriksen, Ian and Quintero, E. A. and
            Harris, Charles R. and Archibald, Anne M. and
            Ribeiro, Ant{\^o}nio H. and Pedregosa, Fabian and
            {van Mulbregt}, Paul and {SciPy 1.0 Contributors}},
  title   = {{{SciPy} 1.0: Fundamental Algorithms for Scientific
            Computing in Python}},
  journal = {Nat. Methods},
  year    = {2020},
  volume  = {17},
  pages   = {261--272},
  adsurl  = {https://rdcu.be/b08Wh},
  doi     = {10.1038/s41592-019-0686-2},
}

@inproceedings{irf,
  title={Improving the training of rectified flows},
  author={Lee, Sangyun and Lin, Zinan and Fanti, Giulia},
  booktitle={NeurIPS},
  year={2024}
}

@inproceedings{kim2024srgb,
  title={sRGB Real Noise Modeling via Noise-Aware Sampling with Normalizing Flows},
  author={Kim, Dongjin and Jung, Donggoo and Baik, Sungyong and Kim, Tae Hyun},
  booktitle={ICLR},
  year={2024}
}

@inproceedings{pixel_shuffle,
  title={Real-time single image and video super-resolution using an efficient sub-pixel convolutional neural network},
  author={Shi, Wenzhe and Caballero, Jose and Husz{\'a}r, Ferenc and Totz, Johannes and Aitken, Andrew P and Bishop, Rob and Rueckert, Daniel and Wang, Zehan},
  booktitle={CVPR},
  year={2016}
}

@inproceedings{adam,
  title={Adam: A method for stochastic optimization},
  author={Kingma, Diederik P},
  booktitle={ICLR},
  year={2015}
}

@inproceedings{idf,
  title={IDF: Iterative dynamic filtering networks for generalizable image denoising},
  author={Kim, Dongjin and Ko, Jaekyun and Ali, Muhammad Kashif and Kim, Tae Hyun},
  booktitle={ICCV},
  year={2025}
}

\clearpage
\setcounter{page}{1}
\maketitlesupplementary

% reset numbering
\setcounter{figure}{0}
\setcounter{table}{0}
\setcounter{equation}{0}
\setcounter{section}{0}

\renewcommand{\thesection}{S\arabic{section}}
\renewcommand{\thefigure}{S\arabic{figure}}
\renewcommand{\thetable}{S\arabic{table}}
\renewcommand{\theequation}{S\arabic{equation}}
% \renewcommand{\thealgocf}{A\arabic{algocf}}

% package
\definecolor{red}{RGB}{220, 0, 0} 
\definecolor{blue}{RGB}{0, 60, 200}

\section{Implementation Details}
\subsection{Fixed-Scale SR.}  
Using our synthetic LR–HR pairs, we train five representative SR models: RCAN~\cite{rcan}, HAN~\cite{han}, SwinIR~\cite{swinir}, HAT~\cite{hat}, and MambaIR~\cite{mambair}.  
Following InterFlow~\cite{interflow}, we spatially downsample LR inputs via the pixel-unshuffle operation~\cite{pixel_shuffle} to align their spatial dimensions with the input size of the SR network, ensuring a fair comparison.
All models are trained with an $\mathcal{L}_1$ loss and optimized using Adam~\cite{adam}, with an initial learning rate of $1\times10^{-4}$ that is halved at 50\%, 75\%, and 87.5\% of 300k iterations. Training is performed using 16 randomly cropped $128\times128$ patches per mini-batch, with random flips for data augmentation.

\subsection{Arbitrary-Scale SR.}  
Following InterFlow’s training settings, we train three SR models: MetaSR~\cite{metasr}, LIIF~\cite{liif}, and CiaoSR~\cite{ciaosr}.
Since~\framework{} synthesizes LR images at the same resolution as their HR counterparts, 
we follow InterFlow~\cite{interflow} and apply bicubic downsampling with scale factor $s$ to the synthesized LR images from~\framework{} for arbitrary-scale SR training. 
While this additional downsampling may introduce additional degradations, 
we empirically observe that our method still achieves superior downstream SR performance when trained on the synthesized datasets. 
We leave further investigation of this constraint for future work.

All models are trained with an $\mathcal{L}_1$ loss and optimized using Adam ($1\times10^{-4}$ initial learning rate, halved at 50\%, 75\%, and 87.5\% of 200k iterations).
Training uses 8 cropped patches per mini-batch with random flips.

\subsection{Evaluation Metrics}
To assess the quality of the synthesized degradations, 
we use PSNR to measure fidelity with respect to the target degradation level, 
CLIP~\cite{clip} to evaluate semantic consistency between synthesized and real degradations, 
and FID~\cite{fid} to measure the perceptual realism of the generated degradations.

To evaluate the performance of SR models, we employ PSNR and SSIM~\cite{ssim} to measure reconstruction fidelity, and LPIPS~\cite{lpips} to assess perceptual quality.

\subsection{LFM Inference}
To obtain the latent representation at an arbitrary target degradation level, 
as illustrated in Fig.~2(b) of the main manuscript, 
we numerically integrate the learned velocity field $\hat{z}'_t$ along the predicted trajectory (see Eq.~(2) in the main manuscript) 
using the adaptive Runge-Kutta (RK45) ODE solver~\cite{rk45_1, rk45_2} from SciPy~\cite{scipy}. 
The integration begins from $z_{t=0}$ and proceeds until the desired timestep $t$, 
which corresponds to the target degradation level. 
The RK45 solver adaptively adjusts the integration step size, 
enabling efficient computation of latent states 
while minimizing the number of function evaluations (NFEs).

\section{Model Size and Inference Speed}

All experiments are conducted on an NVIDIA RTX A6000 GPU for both training and evaluation. 
The DegFlow framework consists of three main components: the RAE encoder (14.6M parameters), 
the RAE decoder (14.8M parameters), and the NCSN++ backbone~\cite{song_score} for LFM (60.5M parameters), 
resulting in a total of approximately 90M parameters.

We evaluate the average inference speed on the RealSR Canon test set, 
whose images have an average resolution of $1300 \times 1040$. 
For different timesteps $t$, which correspond to different target degradation scales, 
we report the number of function evaluations (NFE) using the RK45 ODE solver and the corresponding runtime per image:
1.3 s with NFE=33.08 for $s=2$, 
2.1 s with NFE=49.52 for $s=3$, 
and 2.8 s with NFE=67.40 for $s=4$.

The runtime scales proportionally with the integration length in the latent space, 
while remaining within a practical range for real-world SR dataset generation.

\section{Additional Ablation Studies}

\begin{figure}[!t]
\begin{center}
\centerline{\includegraphics[width=1.0\columnwidth]{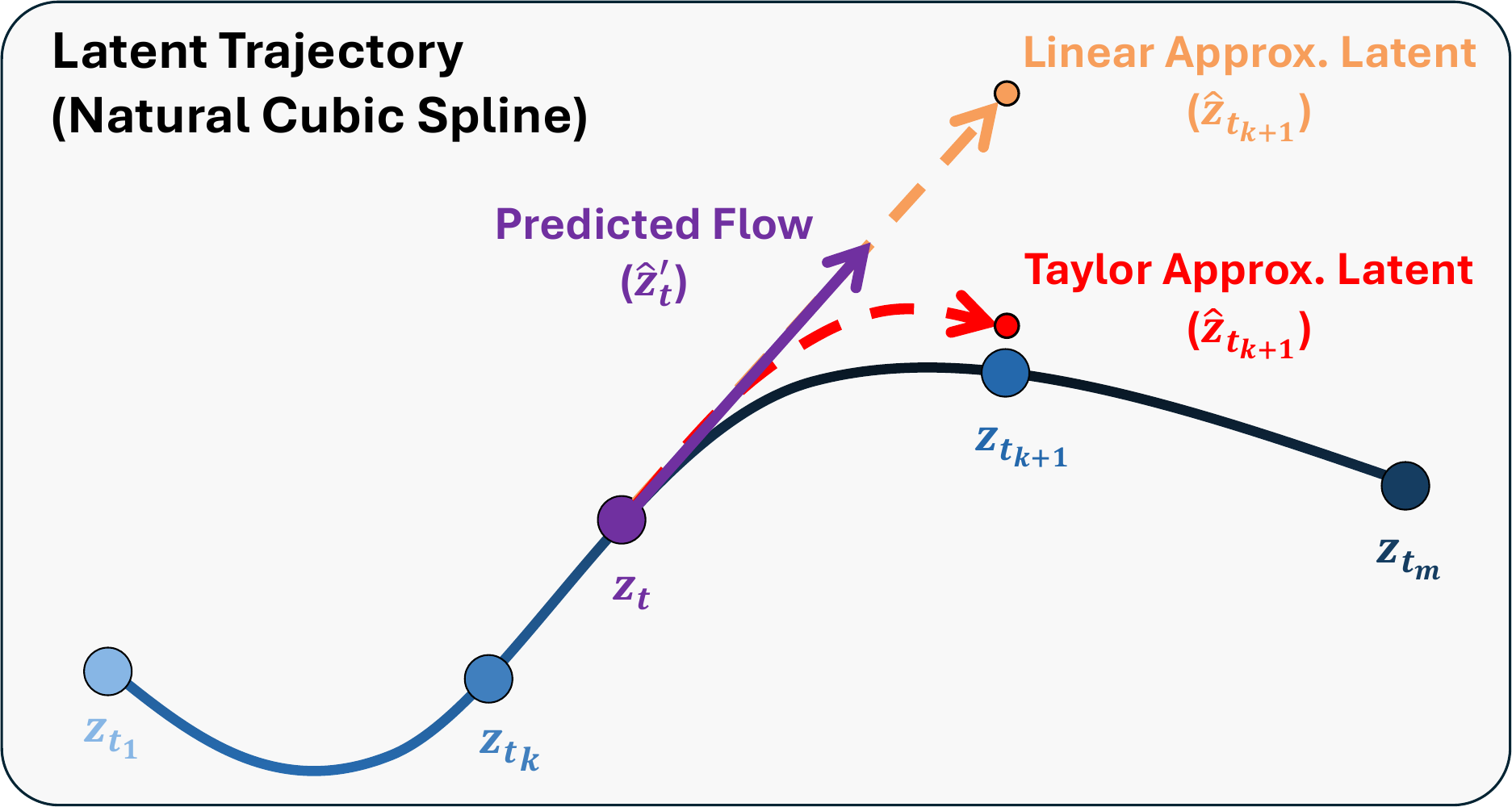}}
% \vspace{-10px}
\caption{Illustration of latent extrapolation from an intermediate timestep $t_k\leq t \leq t_{k+1}$ toward the next degradation level $t_{k+1}$ along the natural cubic spline trajectory.}
\label{fig:lpip_approx}
\vspace{-5mm}
\end{center}
\end{figure}

\begin{table}[!t]
\centering
\newcommand{\red}[1]{\textcolor{red}{\textbf{#1}}}
\caption{Effect of different perceptual loss variants on DegFlow when used to synthesize the SR training dataset ($\times2$\ssim$\times4$) for the HAT network.}
\resizebox{0.8\columnwidth}{!}{
\begin{tabular}{l!{\vrule width 0.4pt}ccc}
\toprule
\rowcolor{HeaderBlue}
\multirow{2}{*}{} & \multicolumn{3}{c}{RealSR $\times3$ Test Set} \\
\rowcolor{HeaderBlue}
\multirow{-2}{*}{\shortstack[l]{Perceptual Loss\\ Variants}}
 & PSNR$\uparrow$ & SSIM$\uparrow$ & LPIPS$\downarrow$ \\ \midrule
w/o LPIPS               & 30.74 & 0.8655 & 0.3206 \\
LPIPS (Linear Approx.)  & 30.60 & 0.8651 & 0.3220 \\
\cellcolor{gray!10}LPIPS (Taylor Approx.)  & \cellcolor{gray!10}\red{30.81} & \cellcolor{gray!10}\red{0.8662} & \cellcolor{gray!10}\red{0.3200} \\
\bottomrule
\end{tabular}
}
\label{tab:lpip_approx}
\end{table}

\subsection{Effect of Perceptual Loss Variants.}
We evaluate the impact of different perceptual loss variants on DegFlow when generating the synthetic SR training dataset ($\times 2$\ssim$\times 4$) for the HAT network.
As illustrated in~\ffigref{fig:lpip_approx} and described in Eq.~(11) of the main manuscript, 
given a predicted flow $\hat{z}'_t$ at an intermediate timestep $t$, the target latent at the next degradation level $t_{k+1}$ must be estimated in order to compute the LPIPS loss with respect to the corresponding ground-truth LR image.

A straightforward approach, as proposed by~\cite{irf}, is to use a linear approximation:
\begin{equation}
\hat z_{t_{k+1}}
 = z_t + \hat z'_t\,\Delta t ,
\label{eq:linear}
\end{equation}
where $\Delta t = t_{k+1}- t$. This directly projects the velocity $\hat{z}'_t$ toward $t_{k+1}$. However, in our case, the probability path is parameterized by a natural cubic spline, and thus this linear constraint is suboptimal as it ignores the nonlinear curvature of the trajectory.

Instead, we employ a third-order Taylor approximation that is inherently consistent with the cubic nature of the predefined trajectory in Eq. (11) of the main manuscript.
This formulation more accurately follows the curvature of the natural cubic spline, ensuring that the extrapolated latent remains faithful to the true trajectory geometry.

\ttabref{tab:lpip_approx} reports the SR results on the RealSR $\times 3$ test set using datasets synthesized from each variant.
The Taylor Approximation achieves the best overall performance (PSNR: 30.81 dB, SSIM: 0.8662, and LPIPS: 0.3200), surpassing both the baseline without (w/o) LPIPS and the linear-approximation variant.
These results confirm that aligning the approximation order with the order of the predefined latent trajectory produces more accurate intermediate degradations, which in turn enhances SR performance.

\begin{table}[]
\centering
\newcommand{\red}[1]{\textcolor{red}{\textbf{#1}}}
\caption{Effect of incorporating RAE’s HR feature skip connections on the NFE required by the LFM when using the RK45 ODE sampler. NFEs are measured on the RealSR ×3 test set at different timesteps $t$, corresponding to approximate degradation levels $s \approx 2, 3, 4$.}
\resizebox{1.0\columnwidth}{!}{
\begin{tabular}{c!{\vrule width 0.4pt}c!{\vrule width 0.4pt}c!{\vrule width 0.4pt}c}
\toprule
\rowcolor{HeaderBlue}
\multirow{2}{*}{} & \multicolumn{3}{c}{NFE using RK45}              \\
\rowcolor{HeaderBlue}
\multirow{-2}{*}{\shortstack[l]{RAE's HR Features\\Skip Connection}}
 & $t=\frac{1}{3}$  (s$\approx$2) & $t=\frac{2}{3}$ (s$\approx$3)  & $t=1$ (s$\approx$4)     \\ \midrule
\xm                                                                                             & 41.36          & 60.68          & 99.2          \\
\cellcolor{gray!10}\cm                                                    & \cellcolor{gray!10}\red{33.08} & \cellcolor{gray!10}\red{49.52} & \cellcolor{gray!10}\red{67.4} \\ \bottomrule
\end{tabular}
}
\label{tab:lfm_sampling}
\end{table}

\subsection{Impact of HR Feature Skip Connections on LFM Efficiency}
We investigate the effect of incorporating HR feature skip connections in the RAE on the efficiency of the LFM model. 
These skip connections connect multi-scale HR features directly to the decoder, allowing the latent representation to concentrate primarily on degradation-specific residual information. 

~\ttabref{tab:lfm_sampling} shows the number of function evaluations (NFE) required by the LFM when using the RK45 sampler~\cite{rk45_1, rk45_2} 
at different timesteps on the RealSR $\times$3 test set. 
Without skip connections, the LFM must encode both low- and high-frequency information in the latent, 
which increases the integration complexity and leads to significantly higher NFEs (\eg 60.68 vs. 49.52 at $t = \frac{2}{3}$, $s \approx 3$).

In contrast, with skip connections, degradation-related residuals are more cleanly embedded in the latent representation, reducing integration complexity and lowering NFEs consistently across all timesteps. 
These results confirm that HR feature skip connections not only enhance reconstruction fidelity but also improve the sampling efficiency of the LFM.

\begin{figure}[!t]
\begin{center}
\centerline{\includegraphics[width=1.0\columnwidth]{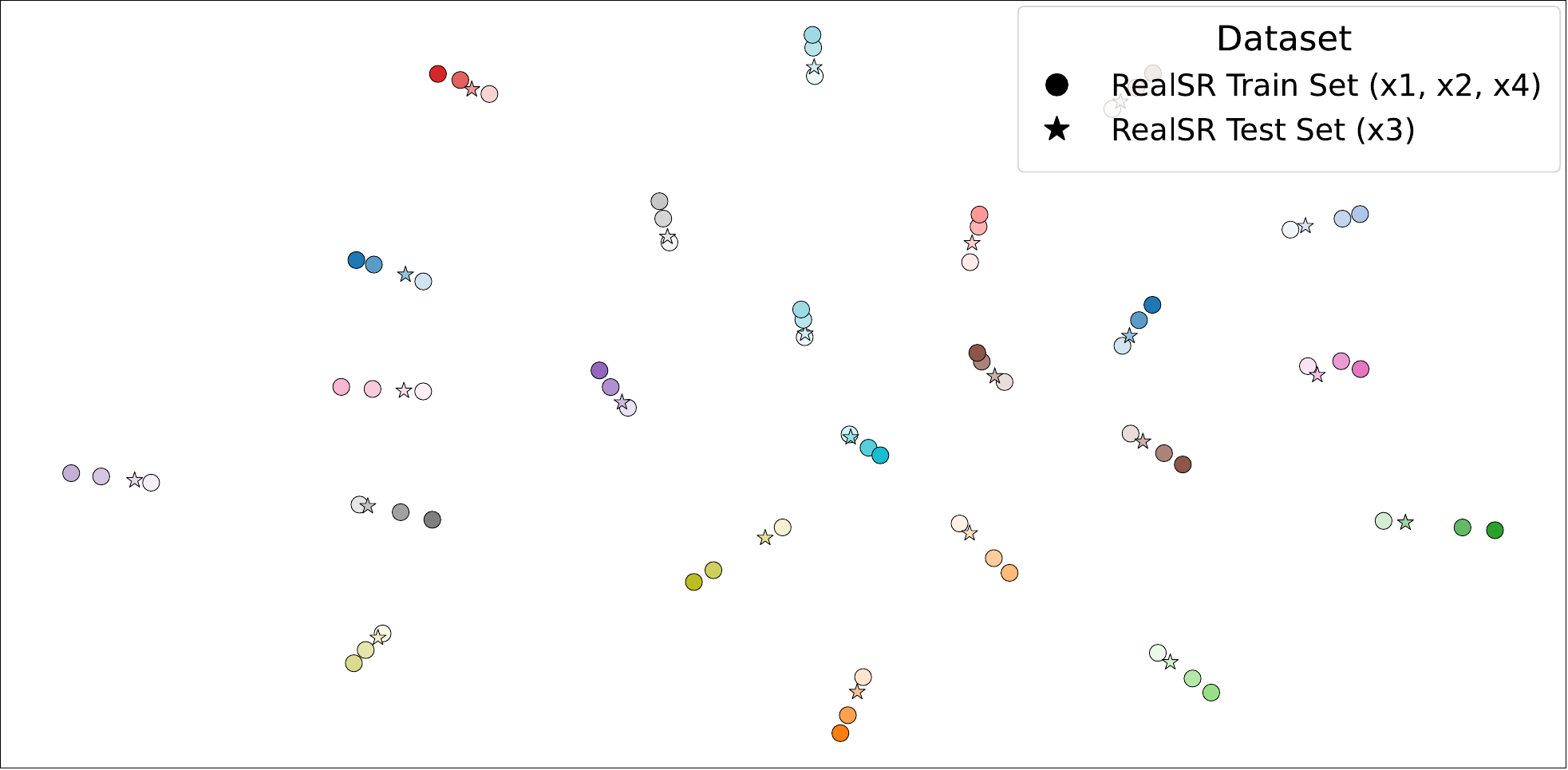}}
% \vspace{-10px}
\caption{t-SNE visualization of RAE latents extracted from the RealSR dataset. The training set consists of scales $\times$1, $\times$2, and $\times$4 (circles), while the test set consists of scale $\times$3 (stars). Points with similar colors correspond to the same image content, and different gradations of the same color indicate different degradation scales (darker = lower degradation level, lighter = higher degradation level).}
\label{fig:latent_vis}
\vspace{-5mm}
\end{center}
\end{figure}

\begin{figure*}[!ht]
\begin{center}
\centerline{\includegraphics[width=0.95\textwidth]{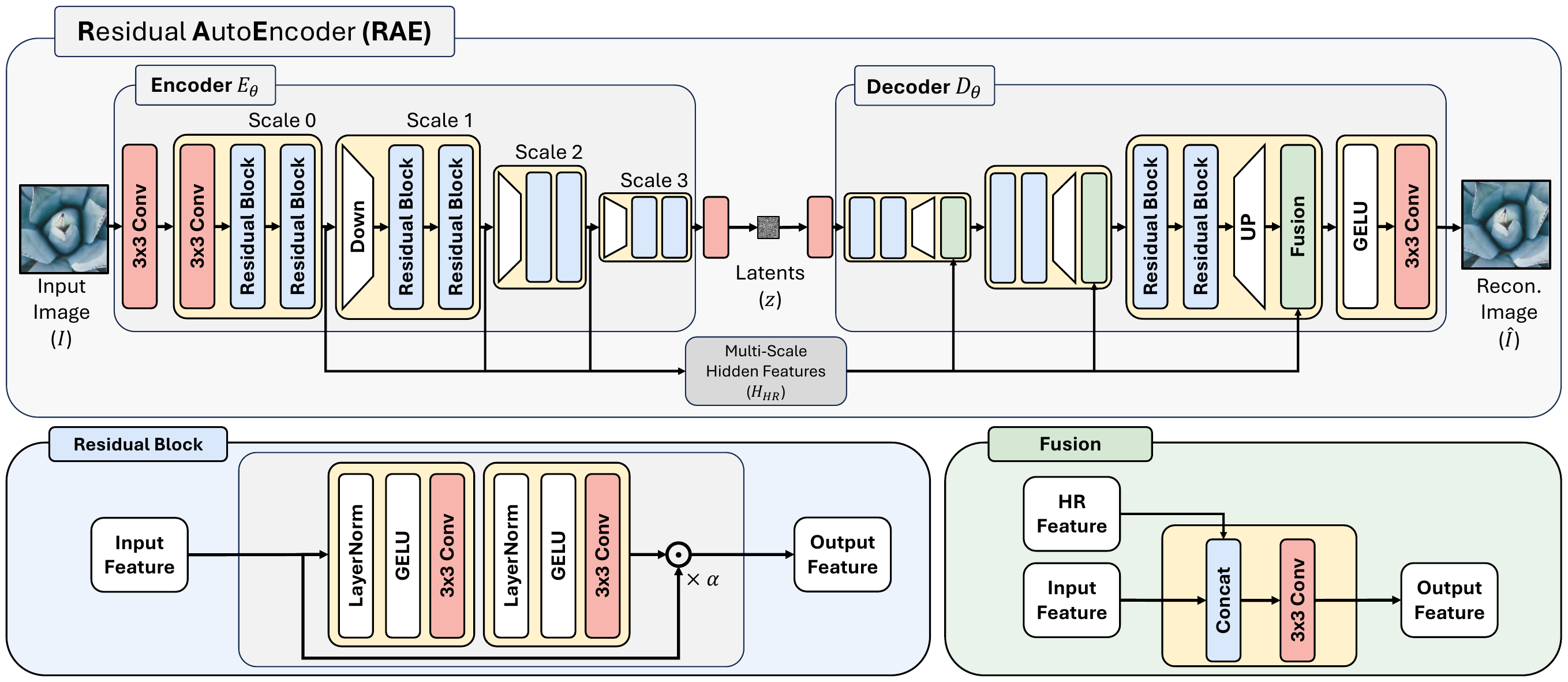}}
\caption{Overview of the RAE Architecture.}
\label{fig:rae_overall_flow}
\vspace*{-10mm}
\end{center}
\end{figure*}

\subsection{Visualization of Degradation Representation in RAE Latent Space}
To examine whether the RAE learns a degradation-aware continuous latent space, we visualize its latent representations using t-SNE. 
The RAE is trained on RealSR training images at scales $\times$1, $\times$2, and $\times$4, 
while the unseen scale $\times$3 from the RealSR test set is included for evaluation.

In~\ffigref{fig:latent_vis}, points with the same base color correspond to the same image content, 
while different gradations of the same color represent different degradation scales (darker = lower degradation level, lighter = higher degradation level). 
The smooth transitions between darker and lighter tones and the close clustering of neighboring scales 
indicate that the latent space is continuous across degradation levels. 
This continuity allows LFM to interpolate along smooth latent trajectories to synthesize intermediate degradations, even for scales not seen during training. 
Such a property is essential for~\framework{} to accurately model and generate LR images at arbitrary degradation levels for real-world SR.

\begin{table}[]
\centering
\newcommand{\red}[1]{\textcolor{red}{\textbf{#1}}}
\newcommand{\blue}[1]{\textcolor{blue}{#1}}
\caption{Effect of RAE Scale Levels. Best and second-best are highlighted in \red{red} and \blue{blue}.
}
\resizebox{0.8\columnwidth}{!}{
\begin{tabular}{l!{\vrule width 0.4pt}ccc}
\toprule
\rowcolor{HeaderBlue}
\multirow{2}{*}{} & \multicolumn{3}{c}{RealSR $\times$3 Test Set}             \\ %\cmidrule{2-4}
\rowcolor{HeaderBlue}
\multirow{-2}{*}{\shortstack[l]{RAE \\ \# Scale Levels}}
                                                                                & PSNR↑          & SSIM↑           & LPIPS↓          \\ \midrule
w/o RAE (Pixel Space)                                                            & 30.34          & 0.8564          & 0.3239          \\
2                                                                               & 30.32          & 0.8561          & 0.3225          \\
4                                                                               & \blue{30.66}   & \blue{0.8643}          & \red{0.3115} \\
\cellcolor{gray!10}8 (Ours)                                                      & \cellcolor{gray!10}\red{30.75}    & \cellcolor{gray!10}\red{0.8651} & \cellcolor{gray!10}\blue{0.3194}          \\
16                                                                              & 30.64          & 0.8639          & 0.3220          \\ \bottomrule
\end{tabular}
}
\vspace{-5mm}
\label{tab:rae_scale}
\end{table}

\subsection{Effect of RAE Scale Levels}
We study the effect of the number of spatial downsampling scale levels in the RAE on DegFlow's ability to generate effective training data for downstream SR. 
For each configuration, a dedicated LFM model is trained using RealSR training set $\times$1, $\times$2, and $\times$4, and the resulting DegFlow variant synthesizes intermediate LR images ($\times 2$\ssim$\times 4$) from the RealSR $\times$1 test set. 
These synthetic datasets are then used to train the HAT SR model.

As shown in~\ttabref{tab:rae_scale}, training without an RAE (pixel space) performs worse across all metrics, indicating the importance of degradation-aware latent modeling. 
Using only two scale levels shows minimal performance gains, suggesting insufficient compression for capturing degradation characteristics effectively. 
Increasing the number of scale levels to four or eight significantly improves PSNR and SSIM, with eight levels achieving the best overall fidelity (PSNR: 30.75 dB, SSIM: 0.8651), while four levels achieve the best perceptual quality (LPIPS: 0.3115). However, further increasing to sixteen levels slightly degrades performance, likely due to over-compression and loss of fine degradation cues.

These results show that choosing the appropriate number of RAE downsampling levels is critical 
for balancing latent compactness and degradation information preservation, with eight levels providing the best trade-off for DegFlow’s degradation modeling and downstream SR performance.

\section{Architecture Details}
\subsection{RAE}
The RAE is designed to extract compact, degradation representations while preserving high-frequency information crucial for reconstruction. 

As shown in~\ffigref{fig:rae_overall_flow}, the RAE consists of an encoder $E_\theta$, 
a decoder $D_\theta$, and a skip connection pathway that connects multi-scale HR features directly to the decoder.

The encoder progressively downsamples the input image $I$ through a sequence of convolutional layers and residual blocks.
At each scale, a $3\times3$ convolution followed by multiple residual blocks processes the feature maps before spatial downsampling. 
This process continues across multiple scale levels, producing the latent representation $z$ at the coarsest scale. 
The encoder also stores intermediate HR features $(H_{\mathrm{HR}})$ from each scale for skip connections.

The final latent representation z is produced at the coarsest scale after a bottleneck convolution 
that reduces the feature dimensionality to 4 channels, inspired by the latent-space compression strategy in Latent Stable Diffusion~\cite{sd1}. 
This bottleneck design strongly encourages a compact latent space that focuses on low-frequency information,
which predominantly characterizes real-world degradations.
Before decoding, the latent representation z is re-expanded to its original channel dimension to facilitate processing in the decoder.

The decoder follows the encoder structure, progressively upsampling the latent representation back to the original resolution. 
At each upsampling stage, the corresponding stored HR features are fused with the upsampled features via a Fusion module, 
which concatenates the feature maps and applies a $3\times3$ convolution. 
This skip connection mechanism allows the latent representation to focus on degradation-specific residual information 
while the decoder reconstructs high-frequency details from the HR pathway.

The residual block consists of LayerNorm~\cite{layernorm}, GELU~\cite{gelu} activation, and $3\times3$ convolution layers, 
followed by a residual connection scaled by a learnable parameter $\alpha$. 
This design stabilizes training and enhances feature representation capacity~\cite{nafnet}. 
The Fusion module mitigates the loss of spatial details introduced by downsampling and ensures faithful reconstruction quality.

Finally, a GELU activation and a final $3\times3$ convolution refine the reconstructed image $\hat{I}$. 
This architecture enables the RAE to model a smooth, continuous latent space for degradation information while maintaining high reconstruction fidelity.

\subsection{LFM}
For the LFM, we adopt the NCSN++ architecture~\cite{song_score} with the default CIFAR-10 configuration from the rectified flow official implementation~\cite{rf}. NCSN++ is a widely used score-based generative modeling backbone that has been extensively validated in the literature and is frequently employed as a baseline in subsequent diffusion and flow matching studies~\cite{rf}.
We adapt the NCSN++ architecture for LFM, as it is relatively lightweight compared to more recent large-scale diffusion backbones, offering strong generative capability while remaining computationally efficient for both training and inference.

\section{Limitation}
Similar to other real-world degradation modeling approaches such as DeFlow and InterFlow, DegFlow can synthesize LR images with the same spatial resolution as their HR counterparts. However, training an arbitrary-scale SR model still requires an additional bicubic downsampling step to match the desired scale factor. This resampling may introduce artificial degradations that deviate from real-world characteristics.

Despite this limitation, our empirical results demonstrate that, even under this constraint, multiple SR networks consistently achieve superior downstream SR performance when trained on our synthesized datasets compared to existing alternatives. We attribute this robustness to the high-quality degradation trajectories learned in the latent space, which provide strong and informative supervision for SR training.

\section{Additional Results}

\begin{table}[]
\centering
\newcommand{\red}[1]{\textcolor{red}{\textbf{#1}}}
\newcommand{\blue}[1]{\textcolor{blue}{#1}}
\caption{Arbitrary-scale SR results on  DRealSR test set ($\times 3$).  $^{\dagger}$ denotes SR models trained with both the synthesized RealSR dataset and an additional synthesized dataset using DIV2K. Best and second-best are highlighted in \red{red} and \blue{blue}.
}
\resizebox{1.0\columnwidth}{!}{
\begin{tabular}{l!{\vrule width 0.4pt}l!{\vrule width 0.4pt}l!{\vrule width 0.4pt}ccc}
\toprule
\rowcolor{HeaderBlue}
\multirow{2}{*}{}  & \multirow{2}{*}{} & \multirow{2}{*}{} & \multicolumn{3}{c}{DRealSR $\times$3   Test Set}          \\
\rowcolor{HeaderBlue}
\multirow{-2}{*}{Model}  & \multirow{-2}{*}{Train Set} & \multirow{-2}{*}{LR Generation Method} & PSNR↑          & SSIM↑           & LPIPS↓          \\ \midrule
\multirow{6}{*}{MetaSR} & RealSR $\times$3                  & None (Oracle)                         & 31.14          & 0.8705    & 0.3596    \\ \cdashline{2-6}\noalign{\vskip 0.5ex}
                        & RealSR $\times$1                  & BSRGAN                                & 30.07          & 0.8579          & 0.3831          \\
                        & RealSR $\times$1                  & Real-ESRGAN                           & 29.00          & 0.8489          & 0.3825          \\
                        & RealSR $\times$2, $\times$4              & InterFlow                             & 31.16          & 0.8651          & 0.3559          \\
                        & RealSR $\times$2, $\times$4              & Ours                                  & \blue{31.65}    &\blue{0.8704}  & \blue{0.3509}    \\
                        & RealSR $\times$2, $\times$4              & \cellcolor{gray!10}Ours$^{\dagger}$   & \cellcolor{gray!10}\red{31.88} & \cellcolor{gray!10}\red{0.8723} & \cellcolor{gray!10}\red{0.3484} \\ \midrule
\multirow{6}{*}{LIIF}   & RealSR $\times$3                  & None (Oracle)                         & 31.15          & 0.8705          & 0.3616          \\ \cdashline{2-6}\noalign{\vskip 0.5ex}
                        & RealSR $\times$1                  & BSRGAN                                & 30.21          & 0.8601          & 0.3823          \\
                        & RealSR $\times$1                  & Real-ESRGAN                           & 29.00          & 0.8493          & 0.3827          \\
                        & RealSR $\times$2, $\times$4              & InterFlow                             & 31.15          & 0.8662          & 0.3613          \\
                        & RealSR $\times$2, $\times$4              & Ours                                  & \blue{31.71}    & \blue{0.8721}    & \blue{0.3560}    \\
                        & RealSR $\times$2, $\times$4              & \cellcolor{gray!10}Ours$^{\dagger}$   &  \cellcolor{gray!10}\red{31.89} &  \cellcolor{gray!10}\red{0.8729} &  \cellcolor{gray!10}\red{0.3497} \\ \midrule
\multirow{6}{*}{CiaoSR} & RealSR $\times$3                  & None (Oracle)                         & 31.45          & 0.8719    & 0.3523    \\ \cdashline{2-6}\noalign{\vskip 0.5ex}
                        & RealSR $\times$1                  & BSRGAN                                & 30.54          & 0.8685          & 0.3612          \\
                        & RealSR $\times$1                  & Real-ESRGAN                           & 29.66          & 0.8580          & 0.3750          \\
                        & RealSR $\times$2, $\times$4              & InterFlow                             & 31.23          & 0.8648          & \blue{0.3494}    \\
                        & RealSR $\times$2, $\times$4              & Ours                                  & \blue{31.57}    & \blue{0.8704}        & 0.3499          \\
                        & RealSR $\times$2, $\times$4              & \cellcolor{gray!10}Ours$^{\dagger}$  &  \cellcolor{gray!10}\red{31.81} &  \cellcolor{gray!10}\red{0.8717} & \cellcolor{gray!10} \red{0.3463} \\ \bottomrule
\end{tabular}
}
\label{tab:drealsr_results}
\end{table}

\subsection{Visualization of Continuous Degradation}
In~\ffigref{fig:gen_lr_vis}, we visualize intermediate outputs along the learned latent trajectory for two examples from the RealSR dataset to further illustrate DegFlow’s ability to model continuous real-world degradations. 

In both examples, DegFlow produces smooth and consistent transitions between degradation levels. For instance, in the first example, the bright metallic highlight gradually loses sharpness and detail in a manner consistent with real LR captures. In the second example, the intricate rooftop patterns progressively soften while retaining structural coherence, closely matching the degradation trend observed in real images.

These visualizations confirm that DegFlow effectively learns a scale-continuous degradation manifold, enabling the synthesis of high-quality intermediate degradations that align with real-world characteristics. This property is essential for generating diverse, scale-specific training datasets for both fixed-scale and arbitrary-scale SR tasks.

\subsection{SR Results}
In~\ttabref{tab:drealsr_results}, we further evaluate SR networks trained on DegFlow-synthesized datasets on the challenging DRealSR $\times$3 test set~\cite{drealsr}, which contains diverse scenes and real-world degradations. 
DegFlow is trained on RealSR $\times$1, $\times$2, and $\times$4, and each SR network is trained on a synthesized dataset 
following the same configurations used in Tab. 3 of the main manuscript.

Across three representative arbitrary-scale SR models, including MetaSR, LIIF, and CiaoSR, 
our method consistently achieves the highest or second-highest performance across PSNR, SSIM, and LPIPS. 
Notably, augmenting the synthesized RealSR training set with an additional HR dataset from DIV2K ($^{\dagger}$) 
further improves performance across all metrics, establishing new state-of-the-art results on this benchmark.

These results demonstrate that augmenting SR training datasets with our method enhances robustness on diverse real-world benchmarks, outperforming strong baselines such as BSRGAN, Real-ESRGAN, and InterFlow, even under complex degradation conditions. Furthermore, we find that augmenting training with DegFlow-synthesized datasets generated solely from HR images shows promising results, suggesting the potential to construct large-scale training datasets entirely from HR sources. This offers a clear practical advantage over approaches such as InterFlow, which require paired LR images.

Furthermore, in~\ffigref{fig:qualitative_results} and ~\ffigref{fig:qualitative_results_2}, we present additional qualitative comparisons on the RealSR $\times$3 test dataset to evaluate the fixed-scale SR models (SwinIR, HAN, and MambaIR) and trained with our DegFlow-synthesized datasets versus InterFlow-synthesized datasets.

\begin{figure*}[!t]
\begin{center}
\centerline{\includegraphics[width=1.0\textwidth]{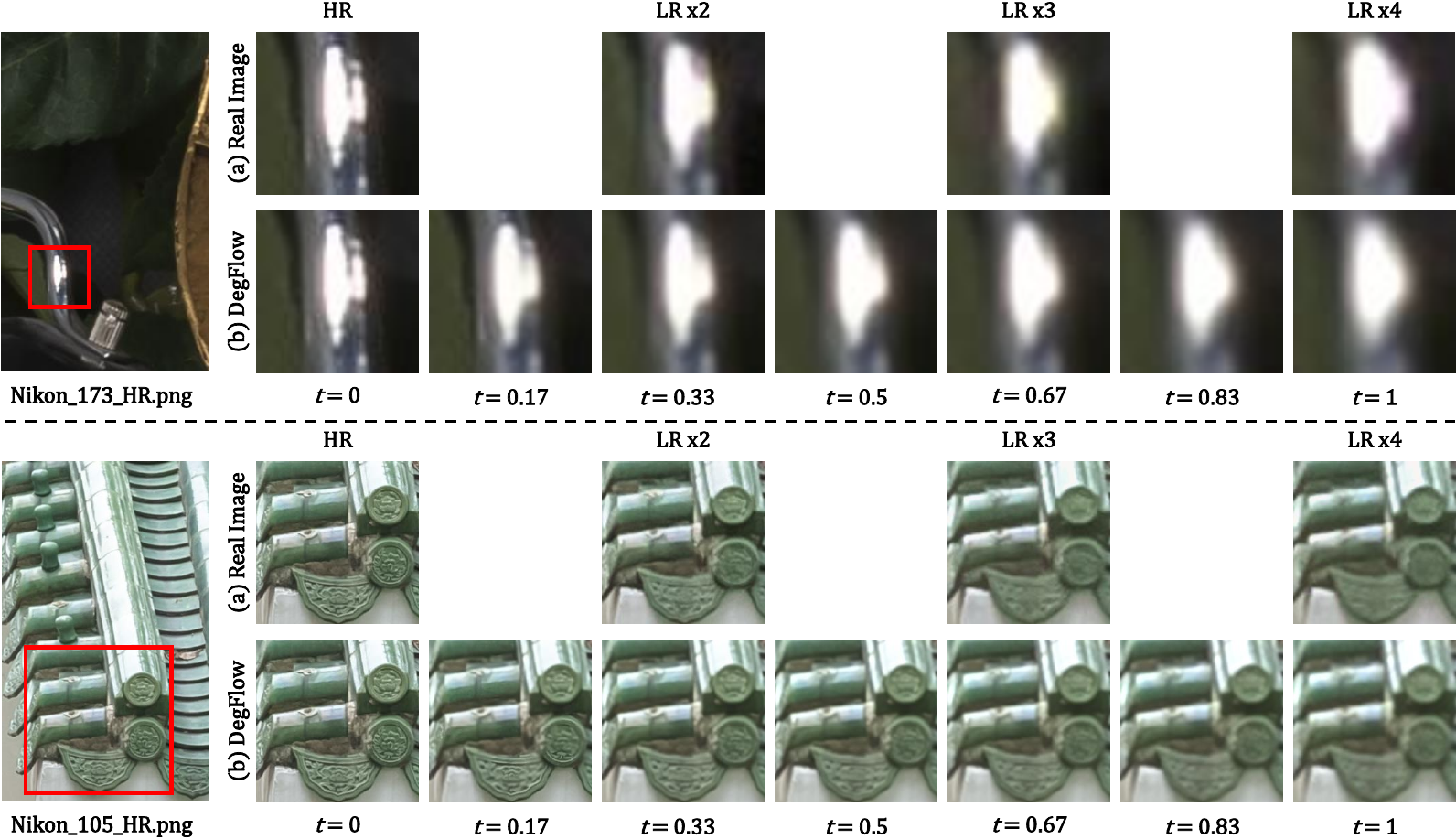}}
% \vspace{-10px}
\caption{Additional visualization of continuous degradation. (a) Real images from the RealSR dataset at discrete scales (HR, ×2, ×3, ×4). (b) \framework{}-generated intermediate degradations at evenly spaced timesteps $0\leq t \leq 1$.
}
\label{fig:gen_lr_vis}
\vspace*{-5mm}
\end{center}
\end{figure*}

\begin{figure*}[!ht]
\begin{center}
\centerline{\includegraphics[width=1.0\textwidth]{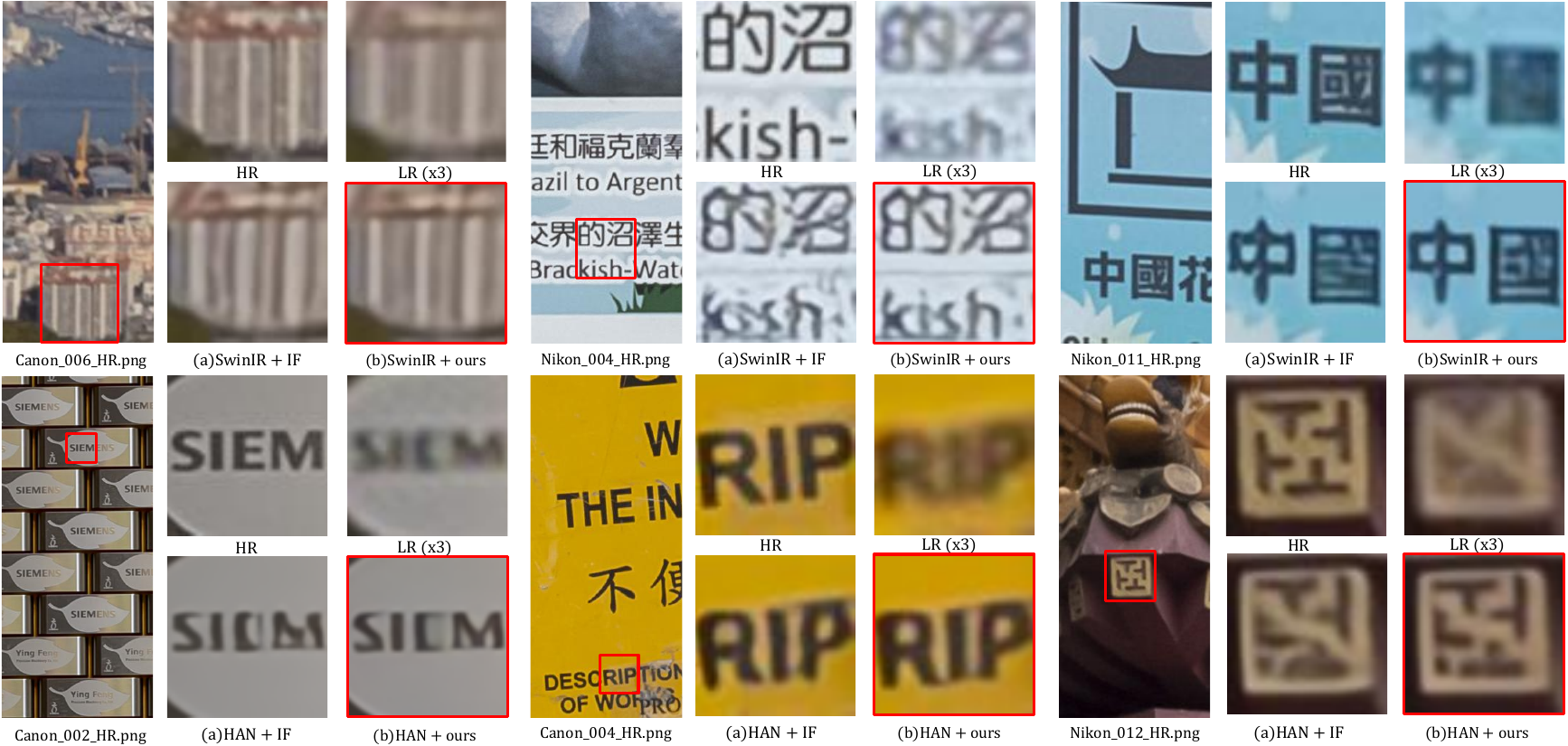}}
% \vspace{-10px}
\caption{Additional qualitative comparisons on the RealSR ×3 dataset. We compare fixed-scale SR results from SwinIR and HAN trained with InterFlow-synthesized datasets (IF) versus those trained with DegFlow-synthesized datasets (Ours). }
\label{fig:qualitative_results}
\vspace*{-10mm}
\end{center}
\end{figure*}

\begin{figure*}[!ht]
\begin{center}
\centerline{\includegraphics[width=1.0\textwidth]{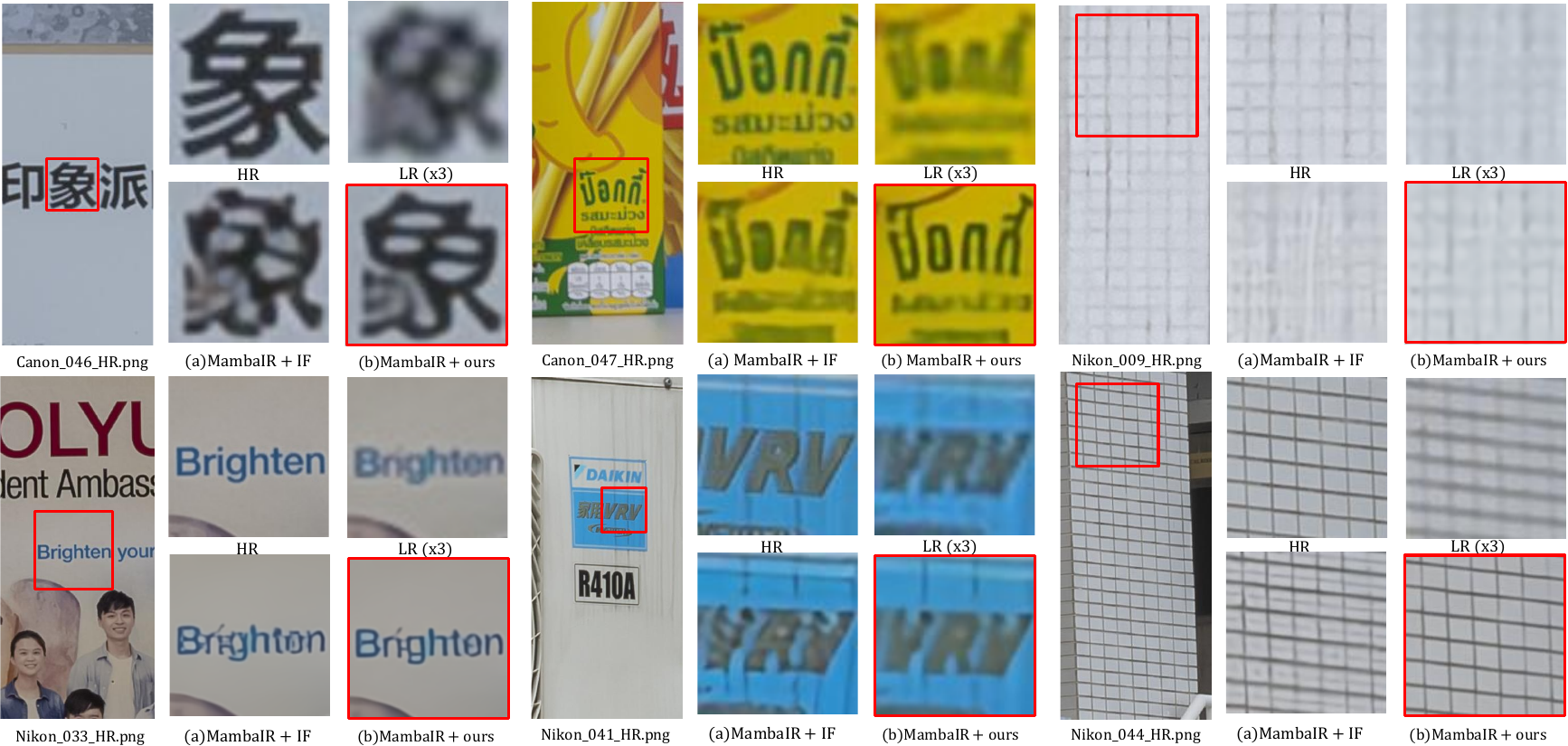}}
% \vspace{-10px}
\caption{Additional qualitative comparisons on the RealSR ×3 dataset. We compare fixed-scale SR results from MambaIR trained with InterFlow-synthesized datasets (IF) versus those trained with DegFlow-synthesized datasets (Ours). }
\label{fig:qualitative_results_2}
\vspace*{-10mm}
\end{center}
\end{figure*}

\clearpage

\end{document}